\documentclass[journal]{IEEEtran}

\usepackage{times}
\usepackage[pdftex]{graphicx}
\usepackage{subfigure}
\usepackage{amsmath,amssymb,amsopn,amstext,amsfonts}
\usepackage{cancel}
\usepackage[space]{cite}
\usepackage{pdfsync}
\usepackage{balance}
\usepackage{color}
\usepackage{mathtools}
\usepackage{bm}

\usepackage{diagbox}
\usepackage{float}
\usepackage{epstopdf}
\usepackage{pifont}
\usepackage{fixltx2e}
\usepackage{amsmath}
\usepackage{multirow}
\usepackage{url}
\usepackage{verbatim}
\usepackage[linkcolor=black,citecolor=black,urlcolor=black,colorlinks=true]{hyperref}
\usepackage{soul,xcolor}
\usepackage[linesnumbered,ruled,vlined]{algorithm2e}
\usepackage{subfiles}
\graphicspath{{../img/}{../../img/}}
\DeclareGraphicsExtensions{,.jpg,.eps,.pdf,.jpeg,.png}

\newcommand{\tnb}[1]{\textnormal{\textbf{#1}}}

\begin{document}

\title{RAPTOR: Robust and Perception-aware Trajectory Replanning for Quadrotor Fast Flight
}
\author{Boyu Zhou,
        Jie Pan,
        Fei Gao,
        and Shaojie Shen
\thanks{Boyu Zhou, Jie Pan and Shaojie Shen are with the Department of Electronic and Computer Engineering, Hong Kong University of Science and Technology, Hong Kong, China. {\tt\footnotesize $\{$bzhouai, jpanai, eeshaojie$\}$@ust.hk}
Fei Gao is with the State Key Laboratory of Industrial Control and Technology, Zhejiang University, Hangzhou 310007, China.{\tt\footnotesize fgaoaa@zju.edu.cn}}%
}


\maketitle

\begin{abstract}
Recent advances in trajectory replanning have enabled quadrotor to navigate autonomously in unknown environments. 
However, high-speed navigation still remains a significant challenge.
Given very limited time, existing methods have no strong guarantee on the feasibility or quality of the solutions. 
Moreover, most methods do not consider environment perception, which is the key bottleneck to fast flight.
In this paper, we present RAPTOR, a robust and perception-aware replanning framework to support fast and safe flight, which addresses these issues systematically.
A path-guided optimization (PGO) approach that incorporates multiple topological paths is devised, to ensure finding feasible and high-quality trajectories in very limited time.
We also introduce a perception-aware planning strategy to actively observe and avoid unknown obstacles.
A risk-aware trajectory refinement ensures that unknown obstacles which may endanger the quadrotor can be observed earlier and avoid in time.
The motion of yaw angle is planned to actively explore the surrounding space that is relevant for safe navigation.
The proposed methods are tested extensively through benchmark comparisons and challenging indoor and outdoor aggressive flights.
We will release our implementation as an open-source package\footnote{To be released at \url{github.com/HKUST-Aerial-Robotics/Fast-Planner}} for the community.
\end{abstract}


\IEEEpeerreviewmaketitle

\vspace{-0.5cm}
\section{Introduction}
\label{sec:intro}

\IEEEPARstart{I}{n} recent years, progresses on different aspects of unmanned aerial vehicles (UAVs), especially quadrotor autonomy have been achieved and promote autonomous navigation. 
Nonetheless, high-speed flight in unknown and highly cluttered environments still remains one of the biggest challenges toward full autonomy.
To achieve fast flight, trajectory replanning is of vital importance to cope with previously unknown obstacles, guaranteeing smooth and safe navigation.

\begin{figure}[t!]
	\begin{center}          
		\subfigure
		{\includegraphics[width=0.49\columnwidth]{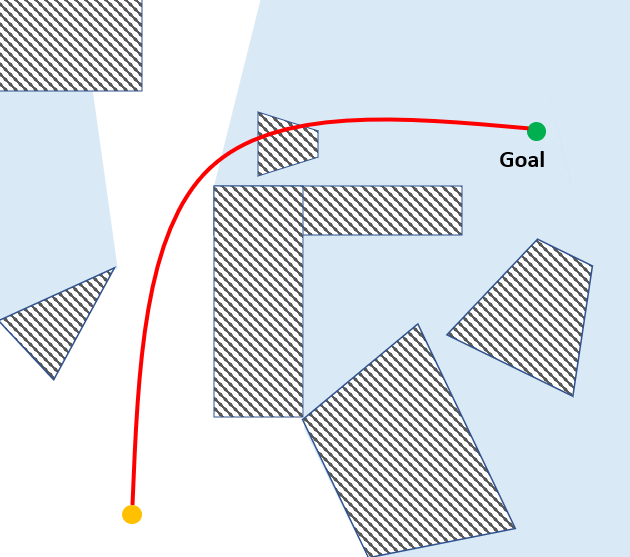}}       
		\subfigure
      {\includegraphics[width=0.49\columnwidth]{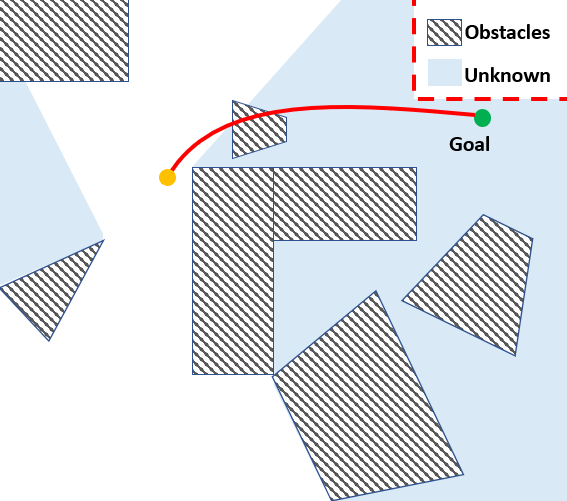}}     
	\end{center}
   \caption{\label{fig:example} An example of planning without perception awareness. 
   The quadrotor flies near the wall, where it has poor visibility to the space behind the corner. 
   In consequence, an obstacle is not revealed until the quadrotor gets very close.
   }
   \vspace{+0.0cm}
\end{figure} 

\begin{figure}[t]
	\begin{center}          
		{\includegraphics[width=0.85\columnwidth]{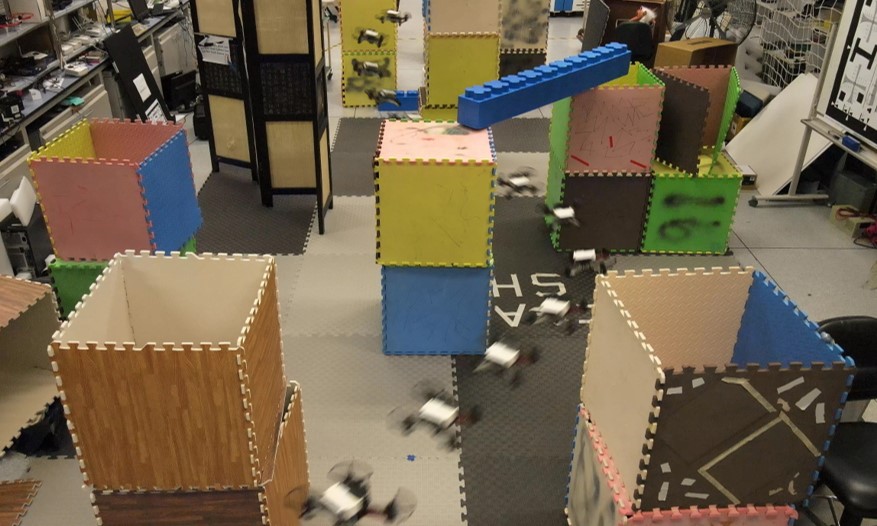}}       
      \vspace{-0.0cm}
	\end{center}
   \caption{\label{fig:flight} Composite image of a fast flight experiment in a cluttered unknown environment.
   Video is available at: \url{https://youtu.be/6wrh4G1-cQ4}.
   }
   \vspace{-0.3cm}
\end{figure} 

Although trajectory replanning has been investigated actively, most presented methods only apply to flights at a moderate speed.
Several issues greatly hinder their usage in high-speed scenarios.
\textbf{(a)} Flying in unknown environments at a high speed, the quadrotor should replan new trajectories to avoid unexpected obstacles in considerably short time, otherwise it crashes.
However, most methods do not guarantee to find feasible trajectories given very limited time.
\textbf{(b)} Current methods typically find a locally optimal trajectory confined within a topologically equivalent class, which does not necessarily contains a satisfactory solution for smooth and safe navigation, especially in fast flight.
\textbf{(c)} Existing methods are unaware of environment perception, which can be fatal as the flight speed and obstacle density get high.
Paying no attention to perception, the planned motions may lead to restricted visibility to the environments, which would in turn result in deficient information of the surrounding space necessary for safe navigation. 
The consequence of not considering perception in replanning can be better illustrated by Fig.\ref{fig:example}. 
To minimize the energy consumption, a trajectory near the wall is generated, along which the visibility toward the unknown space behind the corner is very limited.
As a result the obstacle right behind the corner is invisible until the quadrotor turns right and gets very close, which 'surprises' or even crashes the quadrotor. 
Instead of avoiding what are observed passively, actively observing and avoiding possible dangers is critical for safe high-speed flight.

In this paper, we propose a \textbf{R}obust \textbf{A}nd \textbf{P}erception-aware \textbf{T}raject\textbf{O}ry \textbf{R}eplanning framework called \textbf{RAPTOR} to address these issues systematically. 
To ensure obtaining feasible trajectories within limited time, we present a path-guided gradient-based optimization method, which utilizes geometric guiding paths to eliminate infeasible local minima and guarantee the success of replanning. 
Also, to further improve the optimality of the replanning, we introduce an online topological path planning to extract a comprehensive set of paths that capture the structure of the environment. 
With the guidance of several distinctive paths, multiple trajectories are optimized in parallel, leading to a more thorough exploration of the solution space. 

The above mentioned method that address issues \textbf{(a)} and \textbf{(b)} were first proposed in our previous work \cite{zhou2019robust2}.
However, it adopts the optimistic assumption and lack the awareness of environment perception, which restricts its capability in higher speed and more complex environments.
To bridge this gap, we extend it with a perception-aware planning strategy to enable faster and safer flight from two aspects. 
Firstly, a risk-aware trajectory refinement approach is developed to incorporate with the optimistic planner. 
It identifies unknown regions along the optimistic trajectories that are potentially dangerous to the quadrotor.
Visibility toward such regions along with safe reaction distance are enforced explicitly, ensuring that obstacles in unmapped areas become visible earlier and are avoidable by the quadrotor. 
Secondly, we incorporate the yaw angle of the quadrotor into a two-step motion planning framework. 
A optimal sequence of yaw angles that maximizes information gain and smoothness is searched in the discrete state space, which is further smoothed through optimization. 
The planned motions of yaw angle enable the quadrotor whose field-of-view (FOV) is limited to actively explore the unknown space to gain more relevant knowledge for the future flight.

We conduct systematic evaluations on both the proposed perception-aware planning strategy and the whole planning system, through benchmark comparisons and challenging real-world experiments.
For the former, it is able to support fast and safe flight in challenging scenarios where traditional method fail to ensure safety.
For the later, our planner outperform state-of-the-art methods in several aspects in fast flight tasks.
Extensive indoor and outdoor flight tests in complex environments also validates our planning system.
The contributions of this paper are summarized as follows: 

1) A topological paths-guided gradient-based replanning approach, that is capable of generating high-quality trajectories in limited time.  

2) A risk-aware trajectory refinement approach, which enforces visibility and safe reaction distance to unknown obstacles. 
It improves the predictability and safety of fast flights.  

3) A two-step yaw angle planning method, to actively explore the unknown environments and gather useful information for the flight. 

4) Extensive simulation and real-word tests that validate the proposed method. 
The source code of our system will be released as an open-source package.


\section{Related Work}
\label{sec:related}

\subsection{Quadrotor Trajectory Planning}
\label{subs:related_trajectory}

Trajectory planning for quadrotor has been widely investigated.
Existing methods can be categorized into hard-constrained methods and gradient-based optimization methods.
Hard-constrained methods are pioneered by minimum-snap trajectory\cite{MelKum1105}, in which piecewise polynomial trajectories are generated through quadratic programming(QP). 
\cite{RicBryRoy1312} presented a closed-form solution to minimum snap trajectories.
\cite{CheSuShe2015, fei2018icra, fei2016ssrr, fei2018jfr, ding2019efficient} generate trajectories in a two-step pipeline. 
Safe regions around initial paths are extracted as convex flight corridors, within which QP is solved to generate smooth and safe trajectories. 
Among these methods, poorly chosen time allocation of piecewise polynomials usually lead to unsatisfying results.
To this end, fast marching\cite{fei2018icra} and kinodynamic search\cite{ding2018trajectory, ding2019efficient} are utilized to search for initial paths with more reasonable time allocation.  
\cite{fei2018icra} also proposed to represent trajectories as piecewise B\'ezier curves so that safety and dynamical feasibility are guaranteed.
\cite{tordesillas2019faster} adopted a mixed integer QP formulation to find a more reasonable time allocation of the trajectory.


Another category is the gradient-based trajectory optimization (GTO) methods, which typically formulate trajectory generation as non-linear optimization problems trading off smoothness, safety and dynamic feasibility.
Recently works \cite{oleynikova2016continuous, fei2017iros, lin2018autonomous, usenko2017real, zhou2019robust, zhou2019robust2} revealed that they are particularly effective for local replanning, which is a key component for high-speed flight in unknown environments.
\cite{ratliff2009chomp} proposed to optimize discrete-time trajectory through covariant gradient descent, reviving the community's interest in such methods. 
\cite{kalakrishnan2011stomp} presented a similar formulation, but solves the problem by sampling neighboring candidates iteratively. 
The stochastic sampling strategy partially overcomes the local minima issue but is computationally intensive.
\cite{oleynikova2016continuous} extended it to continuous-time quadrotor trajectories and also adopted random optimization restarts to slightly relieve the typical local minima issue of such methods. 
In \cite{fei2017iros,lin2018autonomous}, the optimization is combined with an informed sampling-based path searching to improve the success rate.
\cite{usenko2017real} proposed to parameterize trajectories as uniform B-splines, showing the usefulness of continuity and locality properties of B-spline for replanning.
However, due to insufficient success rate and efficiency, \cite{oleynikova2016continuous, fei2017iros, usenko2017real} only apply to flight at a moderate speed.  
To this end, \cite{zhou2019robust} further exploited B-spline to improve the efficiency and robustness.
GTO methods are preferable for replanning due to high efficiency.
However, their local minima issue may lead to undesired solutions.
Our method lies in this category, and we resolve local minima by introducing topologically distinctive paths to guide optimization in parallel.

\begin{figure*}[t!]
	\begin{center}          
		{\includegraphics[width=1.9\columnwidth]{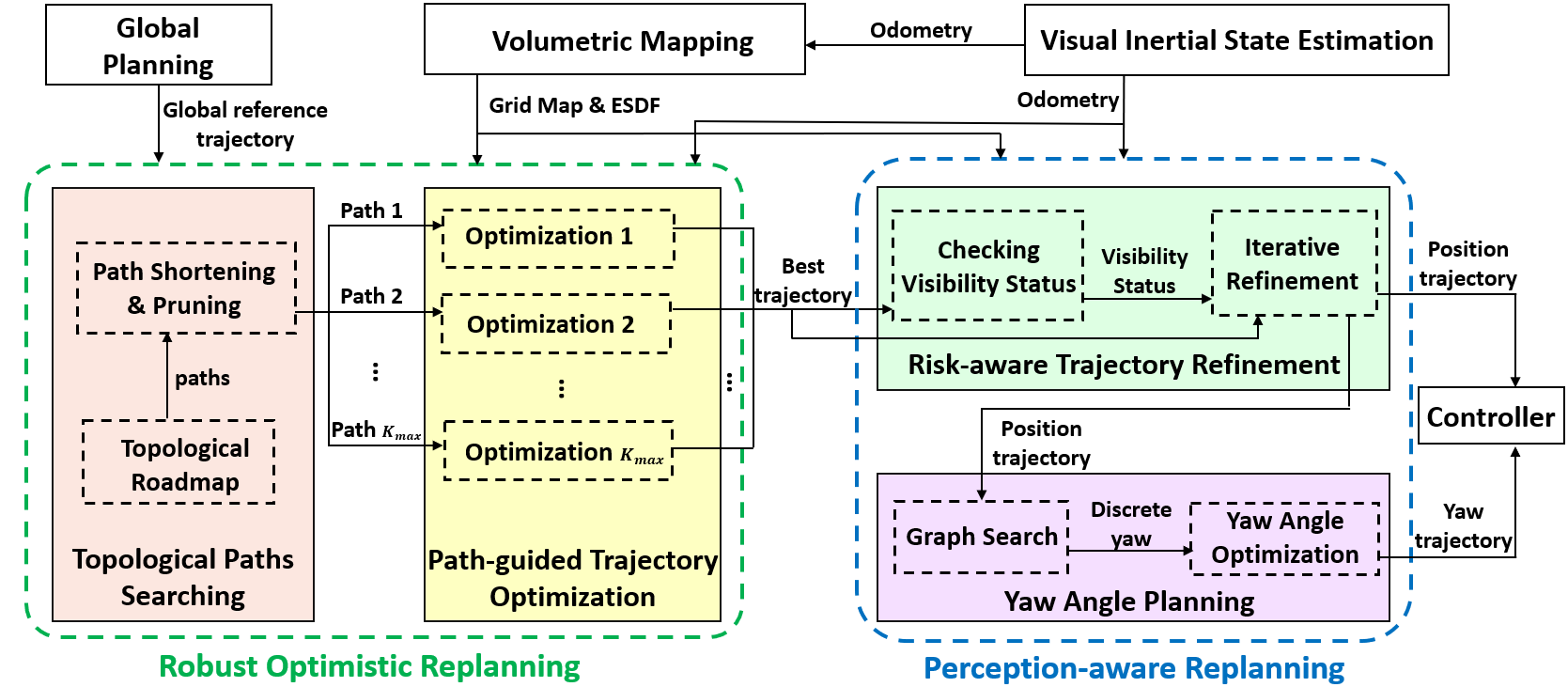}}       
	\end{center}
   \caption{\label{fig:overview} 
   An overview of our replanning system. It takes information from the global planning, mapping and state estimation and replans trajectory in two steps.
   }
\end{figure*} 

\subsection{Topological Path Planning}

There are works utilizing the idea of topologically distinct paths for planning, in which paths belonging to different homotopy (homology) \cite{schmitzberger2002capture,rosmann2015planning,rosmann2017integrated, bhattacharya2010search, bhattacharya2012topological}
or visibility deformation \cite{jaillet2008path} classes are sought. 
\cite{schmitzberger2002capture} constructs a variant of probabilistic roadmap (PRM) to capture homotopy classes, in which path searching and redundant path filtering are conducted simultaneously. 
In contrast, \cite{rosmann2015planning,rosmann2017integrated} firstly creates a PRM or Voronoi diagram, 
after which a homology equivalence relation based on complex analysis\cite{bhattacharya2010search} is adopted to filter out redundant paths. 
These methods only apply to 2D scenarios. 
To seek for 3D homology classes, \cite{bhattacharya2012topological} exploit the theory of electromagnetism and propose a 3D homology equivalence relation. 
However, it requires occupied space to be decomposed into "genus 1" obstacles, which is usually impractical.
Besides, capturing only homotopy classes in 3D space is insufficient to encode the set of useful paths, as indicated in \cite{jaillet2008path}, since 3D homotopic paths may be too hard to deform into each other.
To this end, \cite{jaillet2008path} leverages a visibility deformation roadmap to search for a richer set of useful paths. 
\cite{oleynikova2018sparse, blochliger2018topomap} convert maps built from SLAM systems into sparse graphs representing the topological structure of the environments.
\cite{jaillet2008path, oleynikova2018sparse, blochliger2018topomap} focus on global offline planning and is too time-consuming for online usage. 
Our topological path searching is conceptually closest to \cite{jaillet2008path}, but with a reinvented algorithm for real-time performance. 

\subsection{Navigation in Unknown Environments}

To deal with unknown environments in navigation, different strategies have been used.
Many methods adopt the optimistic assumption \cite{fei2018jfr,liu2017planning, usenko2017real, zhou2019robust}, which treats the unknown space as collision-free.
This strategy improves the chance of reaching goals, but may not guarantee safety. 
On the contrary, some other methods regard unknown space as unsafe and only allow motions within the known-free space \cite{liu2016high} or sensor FOV \cite{lopez2017aggressive,nieuwenhuisen2019search}.
In \cite{lopez2017aggressive2} the sensor FOV constraint is partially relaxed by choosing safe motion primitives generated in the past. 
Although these restrictions ensure safety, they lead to conservative motion.
Recently \cite{tordesillas2019faster} proposed a strategy that plans in both the known-free and unknown space. 
Instead of being over optimistic about the unknown space, it always maintains back-up trajectories to ensure safety.

The limitation of the above mentioned strategies is the lack of environment perception awareness, which is of significant necessity in fast flight.
Although much attention has been paid to planning with the awareness of localization \cite{costante2016perception, zhang2018perception, murali2019perception} and target tracking \cite{falanga2018pampc, jeon2019online}, less emphasis is put on environment perception.
\cite{richter2016learning} proposed a learned heuristic function to guide the path searching into areas with greater visibility to unknown space, but it may not generalize well to complex 3D environments.
\cite{heiden2017planning} showed an integrated mapping and planning framework for active perception. 
The planner iteratively simulates future measurements after executing specific motions, predicts uncertainty of the map, and minimizes the replanning risk. 
Its main drawback is the prohibited runtime for online usage.
In \cite{oleynikova2018safe}, a local planner is coupled with local exploration to safely navigate a cluttered environment.
However, it conservatively selects intermediate goals within known-free space, which restricts the flight speed.
In this paper, we present a perception-aware strategy to ensures that unknown dangers can be discovered and avoided early.
It guarantees safety and does not lead to conservative behaviors. 


\section{System Overview}
\label{sec:overview}

The proposed replanning system is illustrated in Fig.\ref{fig:overview}.
It takes the outputs of the global planning, dense mapping and state estimation modules, and deforms the global reference trajectory locally to avoid previously unknown obstacles.
The replanning works in two steps.
Firstly, the robust optimistic replanning generates multiple locally optimal trajectories in parallel through the path-guided optimization (Sect.\ref{sec:pgo}).
The optimization is guided by topologically distinctive paths extracted and carefully selected from the topological path searching, which will be detailed in Sect.\ref{sec:topo_path}.
Optimistic assumption is adopted in this step.
Secondly, the perception-aware planning strategy is utilized.
The best trajectory among the locally optimal ones is further polished by a risk-aware trajectory refinement, in which its safety and visibility to the unknown and dangerous space is improved, as presented in Sect.\ref{sec:refinement}.
Based on the refined trajectory, the yaw angle is planned to actively explore the unknown environments (Sect.\ref{sec:yaw}). 
The global planning, mapping, estimation and controller are introduced briefly in Sect.\ref{subsec:implement}.


\begin{figure}[t]
	\begin{center}          
		\subfigure[]
		{\includegraphics[width=0.49\columnwidth]{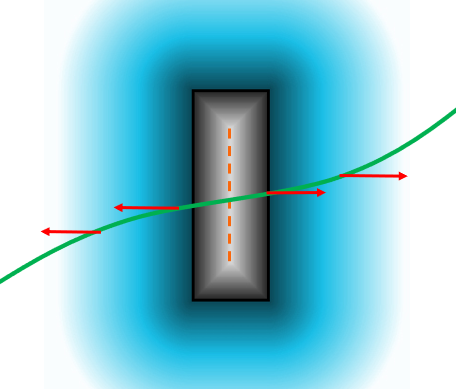}}       
		\subfigure[]
      {\includegraphics[width=0.491\columnwidth]{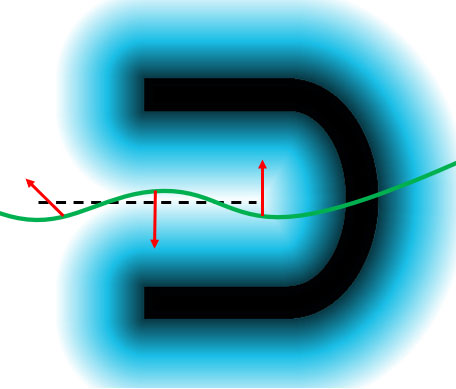}}     
      \vspace{-0.7cm}
	\end{center}
   \caption{\label{fig:failure} Typical examples of optimization failure, where the trajectories pass through the 'I' shape and 'U' shape obstacles.
   (a) Adjacent parts of the trajectory are pushed in opposing directions since it crosses the "valley" of the ESDF (denoted by dashed lines). Negative distance is in gray color, red arrows denote gradients of ESDF. 
   (b) The trajectory crosses the "ridge"  . 
    }
   \vspace{-0.7cm}
\end{figure}  

\section{Path-Guided Trajectory Optimization}
\label{sec:pgo}
As showed in Sect.\ref{subs:related_trajectory}, GTO methods are effective for replanning, but suffer from local minima.
To further improve the robustness of replanning and ensure flight safety, we present PGO, which utilizes a geometric guiding path in the optimization to guarantee its success.

\subsection{Optimization Failure Analysis} 
\label{subs:fail_analy}

Previous work\cite{schulman2014motion} showed that failure of GTO is relevant to unfavorable initialization, i.e., initial paths that pass through obstacles in certain ways usually get stuck.
Underlying reason for this phenomenon is illustrated in Fig.\ref{fig:failure}.
Typical GTO methods incorporate the gradients of a Euclidean signed distance field (ESDF) in a collision cost to push the trajectory out of obstacles.
Yet there are some "valleys" or "ridges" in the ESDF, around which the gradients differ greatly. 
Consequently, if a trajectory is in collision and crosses such regions, the gradients of ESDF will change abruptly at some points.
This can make gradients of the collision cost push different parts of the trajectory in opposing directions and fail the optimization.

Normally, such "valleys" and "ridges", which corresponds to the space that has an identical distance to the surfaces of nearby obstacles, are difficult to avoid, especially in complex environments. 
Therefore, optimization depending solely on the ESDF fails inevitably at times. 
To solve the problem, it is essential to introduce extra information that can produce an objective function whose gradients consistently deform the trajectory to the free space.

\subsection{Problem Formulation}
\label{subs:pro_form}
We propose PGO built upon our previous work \cite{zhou2019robust} that represents trajectories as B-splines for more efficient cost evaluation.
For a trajectory segment in collision, we reparameterize it as a $p_b$ degree uniform B-spline with control points $ \left\{\mathbf{q}_{0}, \mathbf{q}_{1}, \ldots, \mathbf{q}_{N}\right\} $ and knot span $ \Delta t $.

\begin{figure}[t]
	\begin{center}          
		\subfigure[\label{fig:pggto1}]
		{\includegraphics[width=0.49\columnwidth]{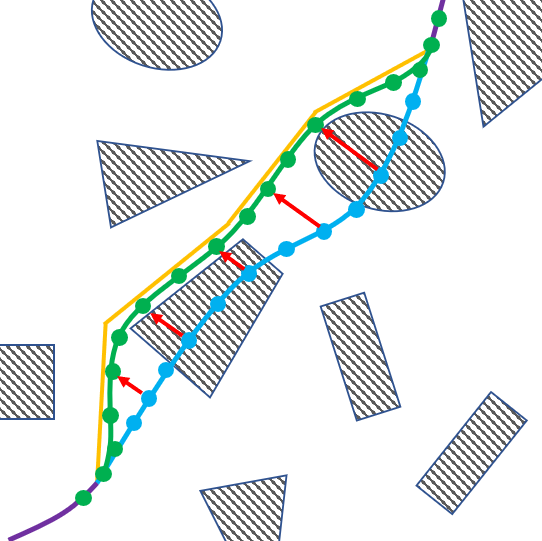}}       
		\subfigure[\label{fig:pggto2}]
		{\includegraphics[width=0.49\columnwidth]{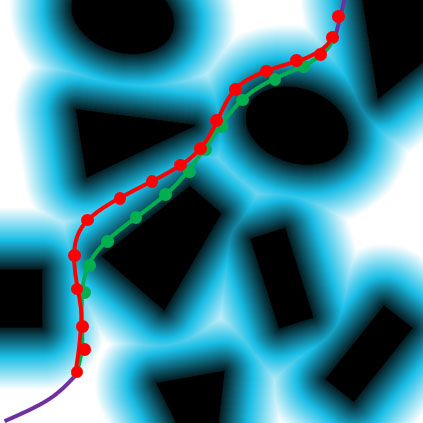}}     
      \vspace{-0.5cm}
	\end{center}
   \caption{\label{fig:pggto} The two-phases PGO approach for trajectory replanning. (a) A geometric guiding path (yellow) attracts the original B-spline trajectory (blue) into the free space.
   (b) The warmup trajectory is further optimized in the ESDF to find a locally optimal trajectory (red).}
   \vspace{-0.5cm}
\end{figure} 

PGO consists of two different phases. The first phase generates an intermediate \textbf{warmup trajectory}.
As concluded above, external information should be included to effectually deform the trajectory, since solely applying the ESDF could be futile.
We employ a geometric guiding path to attract the initial trajectory to the free space (depicted in Fig. \ref{fig:pggto1}) since collision-free paths are readily available from standard methods like A* and RRT*. 
In our work, the paths are provided by the sampling-based topological path searching (Sect. \ref{sec:topo_path}).
The first-phase objective function is:
\begin{align}
   & f_{p1}=\lambda_{s} f_{s}+\lambda_{g} f_{g} = \\ \nonumber
   & \lambda_{s}\sum_{i=p_{b}-1}^{N-p_{b}+1}\left\| \mathbf{q}_{i+1} - 2\mathbf{q}_{i} + \mathbf{q}_{i-1} \right\|^{2} + \lambda_{g}\sum_{i=p_{b}}^{N-p_{b}} \left\| \mathbf{q}_{i} - \mathbf{g}_{i} \right\|^{2}
\end{align}
where $ f_s $ is the smoothness term, while $ f_g $ penalize the distance between the guiding path and the B-spline trajectory.
As in \cite{zhou2019robust}, $ f_s $ is designed as a elastic band cost function\footnote{Only the subset of control points $ \{ \mathbf{q}_{p_b},\mathbf{q}_{p_b+1}, \cdots, \mathbf{q}_{N-p_b}\} $ is optimized 
due to the boundary state constraints of the trajectory. $ \mathbf{q}_{p_b-2}$, $\mathbf{q}_{p_b-1}$, $\mathbf{q}_{N-p_b+1} $ and $ \mathbf{q}_{N-p_b+2} $ are needed to evaluate the smoothness.} that simulates the elastic forces of a sequence of springs.
To simplify the design of $f_g $, we utilize the property that the shape of a B-spline is finely controlled by its control points. 
Each control point $ \mathbf{q}_{i} $ is assigned with an associated point $ \mathbf{g}_{i} $ on the guiding path, which is uniformly sampled along the guiding path.
Then $ f_g $ is defined as the sum of the squared Euclidean distance between these point pairs. 

Notably, minimizing $ f_{p1} $ yields an unconstrained quadratic programming problem, so its optimal solution can be obtained in closed form.
It outputs a smooth trajectory in the vicinity of the guiding path.
Since the path is already collision-free, usually the warmup trajectory is also so.
Even though it is not completely collision-free, its major part will be attracted to the free space.
At this stage, the gradients of ESDF along the trajectory vary smoothly, and the gradients of the objective function push the trajectory to the free space in consistent directions.
Hence, standard GTO methods can be utilized to improve the trajectory.

In the second phase, we adopt our previous B-spline optimization method\cite{zhou2019robust} to further refine the warmup trajectory into a smooth, safe, and dynamically feasible one, whose objective function is:
\begin{align}
\label{equ:fp2}
   & f_{p2} =\lambda_{s} f_{s}+\lambda_{c} f_{c}+ \lambda_{d}\left( f_{v} + f_{a} \right) = \\ \nonumber
         & \lambda_{s}\sum_{i=p_{b}-1}^{N-p_{b}+1}\left\| \mathbf{q}_{i+1}-2\mathbf{q}_{i}+\mathbf{q}_{i-1} \right\|^{2} +  \lambda_{c}\sum\limits_{i=p_b}^{N-p_b} \mathcal{F}(d(\mathbf{q}_{i}), d_{\text{min}}) \\ \nonumber
         & + \lambda_{d}\sum\limits_{\begin{subarray}{c}
            \mu \in \\ \{x,y,z \}
         \end{subarray}} \left\{ \sum\limits_{i=p_b-1}^{N-p_b} \mathcal{F}(v_{\text{max}}^2, \dot{q}_{i,\mu}^2)+\sum\limits_{i=p_b-2}^{N-p_b} \mathcal{F}(a_{\text{max}}^2 , \ddot{q}_{i, \mu}^2) \right\}
\end{align}
Here $ \mathcal{F}() $ is a penalty function for general variables:
\begin{equation}\label{equ:potential}
	\mathcal{F}(x, y) = \left\{
	\begin{array}{cl}
	(x-y)^{2} & x \le y \\
	0 & x > y
	\end{array}
	 \right.
\end{equation}
$ f_c $ is the collision cost that grows rapidly if the trajectory gets closer than $d_{\text{min}}$ to obstacles, where $ d(\mathbf{q}) $ is the distance of point $\mathbf{q}$ in the ESDF. 
$ f_v $ and $ f_a $ penalize infeasible velocity and acceleration, in which $ \dot{\mathbf{q}}_i = \left[ \dot{q}_{i,x}, \dot{q}_{i,y}, \dot{q}_{i,z} \right]^\mathrm{T} $ and $ \ddot{\mathbf{q}}_i = \left[ \ddot{q}_{i,x}, \ddot{q}_{i,y}, \ddot{q}_{i,z} \right]^\mathrm{T} $ are the control points of velocity and acceleration.
They can be evaluated by:
\begin{equation}
   \dot{\mathbf{q}}_{i} = \frac{\mathbf{q}_{i+1} - \mathbf{q}_{i}}{\Delta t}, \ \ 
   \ddot{\mathbf{q}}_{i} = \frac{\mathbf{q}_{i+2} - 2\mathbf{q}_{i+1} + \mathbf{q}_{i}}{\Delta t^{2}}
\end{equation}
$v_{\text{max}}$ and $a_{\text{max}}$ are single-axis maximum velocity and acceleration.
The formulations of $ f_c$, $ f_v$, and $ f_a $ are based on the convex hull property of B-spline, thanks to which it suffices to constrain the control points of the B-spline to ensure safety and dynamic feasibility.
For brevity, we refer the readers to \cite{zhou2019robust} for more details.

Although PGO has one more step of optimization compared with previous methods, it can generate better trajectories within shorter time.
The first-phase takes only negligible time, but generate a warmup trajectory that is easier to be further refined, which improve the overall efficiency. 


\begin{figure}[t]
	\begin{center}          
		\subfigure[\label{fig:definition1} The green, blue and yellow trajectories are equivalent under the definition of homotopy, but represent substantially different motions.]
		{\includegraphics[width=0.49\columnwidth]{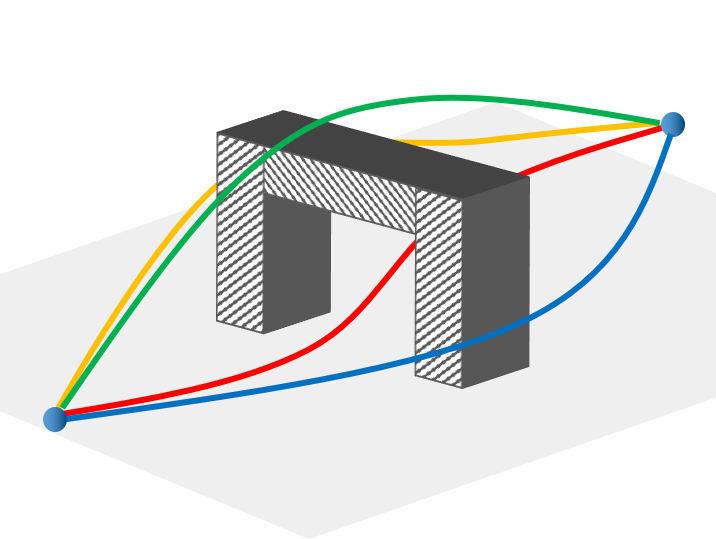}}       
		\subfigure[\label{fig:definition2} An illustration of UVD. The purple trajectory is distinctive to the yellow one, but is equivalent to the blue one.]
		{\includegraphics[width=0.49\columnwidth]{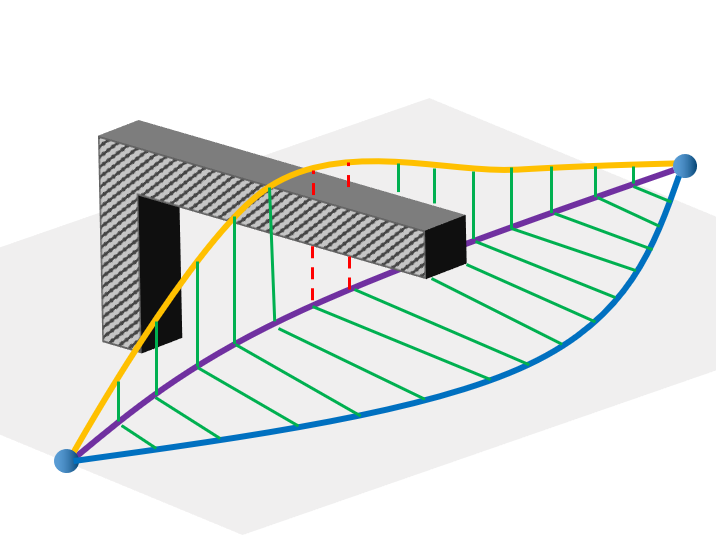}}     
	\end{center}
	\vspace{-0.3cm}
   \caption{\label{fig:definition} Topology equivalence relation.}
   \vspace{-0.8cm}
\end{figure} 

\section{Topological Path Searching}
\label{sec:topo_path}
Given a geometric guiding path, our PGO method can obtain a locally optimal trajectory. 
However, this trajectory is restricted within a topologically equivalent class and not necessarily satisfactory, even with the guidance of the shortest path, as seen in Fig. \ref{fig:roadmap5} and \ref{fig:roadmap6}.
Actually, it is difficult to determine the best geometric path, since the paths do not contain high order information (velocity, acceleration, etc.), and can not completely reflect the true motion.
Searching a kinodynamic path \cite{liu2017iros,webb2013} may suffice, but it takes excessive time to obtain a promising path with boundary state constraints at the start and end of the replanned trajectory.

For a better solution, a variety of guiding paths are required. 
We propose a sampling-based topological path searching to find a collection of distinctive paths. 
Although methods \cite{jaillet2008path,schmitzberger2002capture,oleynikova2018sparse,blochliger2018topomap} are for this problem, none of them runs in real-time in complex 3D environments.
We redesign the algorithm carefully to solve this challenging problem in real time. 

\begin{figure}[t]
	\begin{center}          
		\includegraphics[width=0.99\columnwidth]{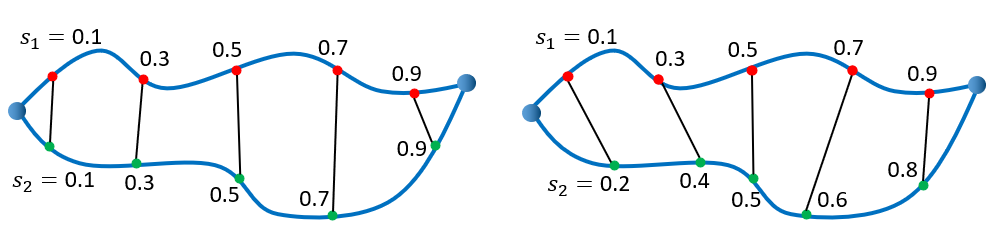}       
	\end{center}
   \vspace{-0.3cm}
   \caption{\label{fig:vduvd} Comparison between UVD (left) and VD (right).
   Each red point on one path is transformed to a green point on the other path.  
   Any two associated points correspond to the same parameter $ s $ in UVD, but not in VD.
   }
   \vspace{-0.7cm}
\end{figure} 

\subsection{Topology Equivalence Relation}
\label{subs:relation}
Although the concept of homotopy is widely used, it captures insufficient useful trajectories in 3D environments, as shown in Fig. \ref{fig:definition1}. 
\cite{jaillet2008path} proposes a more useful relation in 3D space named visibility deformation (VD), but it is computationally expensive for equivalence checking.
Based on VD, we define \textit{uniform visibility deformation} (UVD), which also captures abundant useful trajectories, and is more efficient for equivalence checking.

\begin{algorithm}[t]
   \DontPrintSemicolon
   $ \tnb{Initialize}() $ \;
   $ \tnb{AddGuard}(\mathcal{G}, s), \ \tnb{AddGuard}(\mathcal{G}, g) $ \;
   \While{$ t \le t_{max} \land N_{sample} \le N_{max} $}{
      $ p_s \gets \tnb{Sample}() $ \;
      $ g_{vis} \gets \tnb{VisibleGuards}(\mathcal{G}, p_s) $ \;
      
      \If{$ g_{vis}.\tnb{size}() == 0 $}{
         $ \tnb{AddGuard}(\mathcal{G}, p_s) $ \;
      }
      \If{$ g_{vis}.\tnb{size}() == 2 $}{
         $ path_1 \gets \tnb{Path}(g_{vis}[0], p_s, g_{vis}[1]) $ \;
         $ distinct \gets True $ \;
         $ \mathcal{N}_s \gets \tnb{SharedNeighbors}(\mathcal{G}, g_{vis}[0], g_{vis}[1]) $ \;
         \For{$ \tnb{each} \ n_s \in \mathcal{N}_s $}{
            $ path_2 \gets \tnb{Path}(g_{vis}[0], n_s, g_{vis}[1]) $ \;
            \If{$ \tnb{Equivalent}(path_1, path_2) $}{
               $ distinct \gets False $\;
               \If{$ \tnb{Len}(path_1) < \tnb{Len}(path_2) $}{
                  $ \tnb{Replace}(\mathcal{G}, p_s, n_s) $\;
               }
               break\;
            }
         }
         \If{$ distinct $}{
            $ \tnb{AddConnector}(\mathcal{G}, p_s, g_{vis}[0], g_{vis}[1]) $ \;
         }
      }
   }
   \caption{Topological Roadmap \label{alg:topo_roadmap}}
\end{algorithm}

\textbf{Definition 1.} Two trajectories $ \tau_1(s) $, $ \tau_2(s) $ parameterized by $ s \in [0,1] $ and satisfying $ \tau_1(0) = \tau_2(0), \tau_1(1) = \tau_2(1) $,
belong to the same \textit{uniform visibility deformation} class, if for all $ s $, line $ \overline{\tau_1(s)\tau_2(s)} $ is collision-free.

Fig. \ref{fig:definition2} gives an example of three trajectories belonging to two UVD classes.
The relation between VD and UVD is depicted in Fig. \ref{fig:vduvd}.
Both of them define a continuous map between two paths $\tau_1(s)$ and $ \tau_2(s) $, in which a point on $\tau_1(s)$ is transformed to a point on $\tau_2(s)$ through a straight-line.
The major difference is that for UVD, point $ \tau_1(s_1) $ is mapped to $\tau_2(s_2)$ where $ s_1 = s_2 $, while for VD $ s_1$ does not necessarily equals $ s_2 $.
In concept, UVD is less general and characterizes subsets of VD classes.
Practically, it captures slightly more classes of distinct paths than VD, 
but is far less expensive \footnote{To test VD relation, one should compute a visibility diagram and do path searching within it\cite{jaillet2008path}, which has higher complexity than testing UVD.} for equivalence checking.


To test UVD relation, one can uniformly discretize $ s \in [0,1] $ to $ s_i = i/K, i = 0, 1,..., K $ and check collision for lines $ \overline{\tau_1(s_i)\tau_2(s_i)} $.
For the piece-wise straight line paths (as in Alg. \ref{alg:topo_roadmap}, \textbf{Equivalent}()), we simply parameterize it uniformly, 
so that for any $ s $ except $ \tau(s) $ is the join points of two straight lines, $ \left\| \frac{d\tau(s)}{ds} \right\| = const  $. 

\subsection{Topological Roadmap}
\label{subs:topo_roadmap}

\begin{figure}[t]
	\begin{center}          
		\subfigure[\label{fig:roadmap1}]
		{\includegraphics[width=0.45\columnwidth]{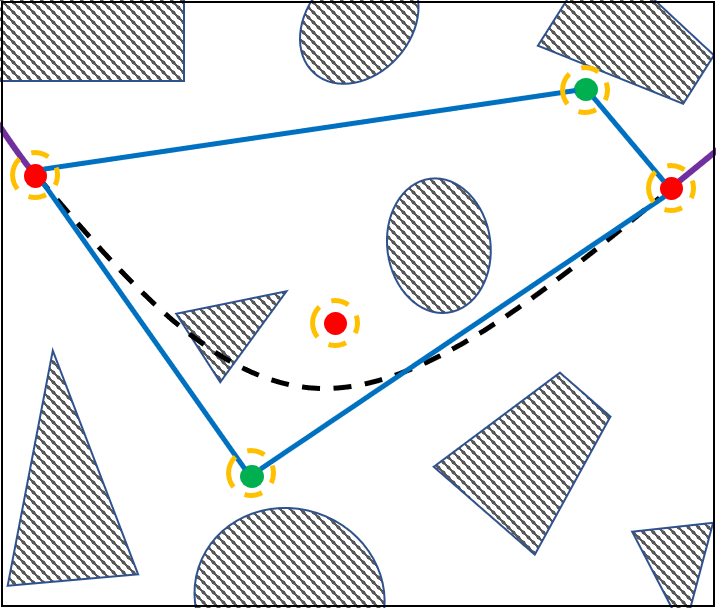}}       
		\subfigure[\label{fig:roadmap2}]
		{\includegraphics[width=0.45\columnwidth]{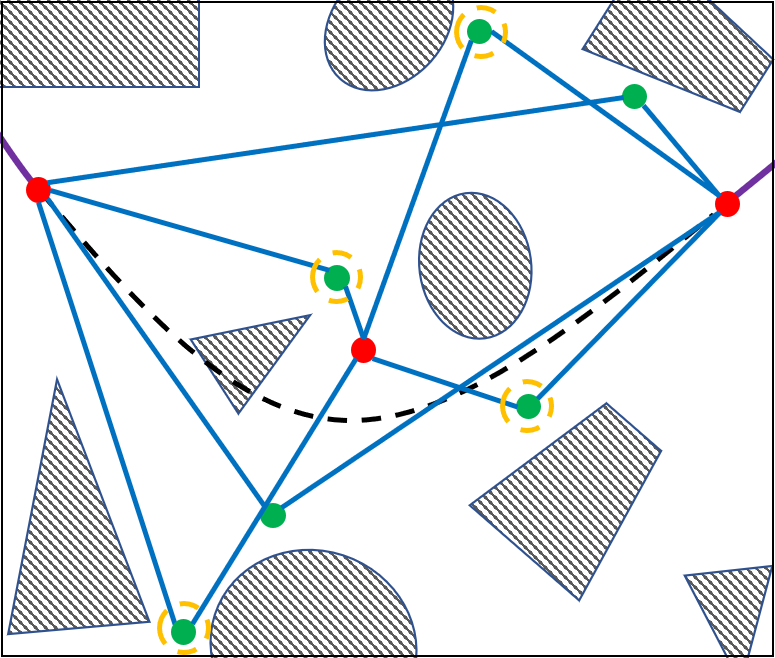}}       
		\subfigure[\label{fig:roadmap3}]
		{\includegraphics[width=0.45\columnwidth]{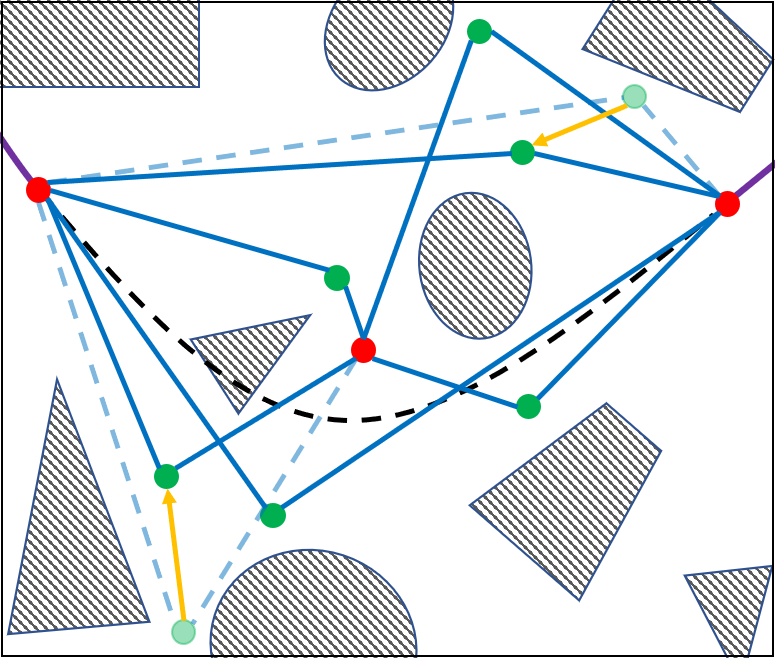}}       
		\subfigure[\label{fig:roadmap4}]
      {\includegraphics[width=0.45\columnwidth]{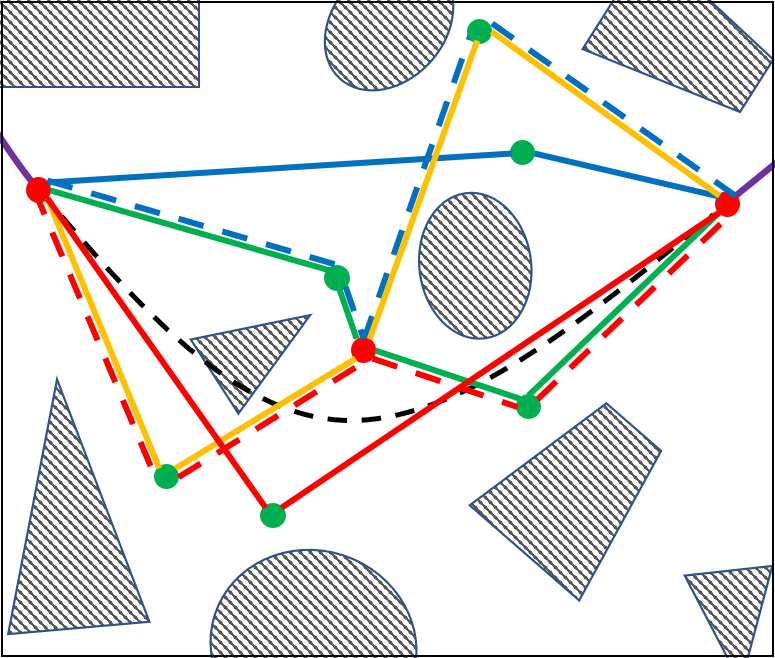}}       
		\subfigure[\label{fig:roadmap5}]
      {\includegraphics[width=0.45\columnwidth]{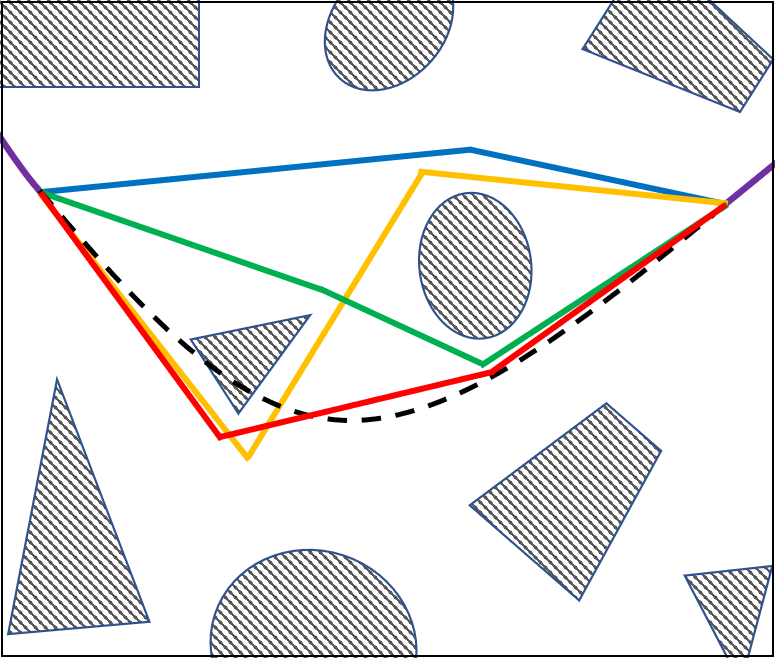}}       
		\subfigure[\label{fig:roadmap6}]
      {\includegraphics[width=0.45\columnwidth]{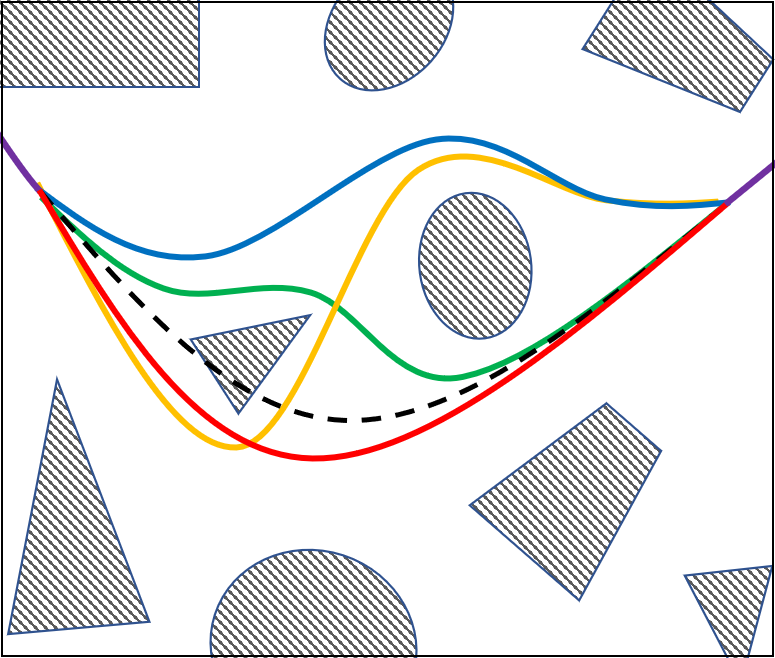}}       
      \vspace{-0.5cm}
	\end{center}
   \caption{\label{fig:roadmap} Topological path searching  and parallel trajectory optimization.
   (a): Two \textit{connectors} (green) are sampled to connect the two initial \textit{guards} (red) at the start and end points, after which a new \textit{guard} is added to occupy a region not covered by others.
   (b): More \textit{connectors} are created to connect the \textit{guards}.
   (c): New \textit{connectors} replace the old ones, making some connections shorter. 
   (d): Paths are extracted from the roadmap. Equivalent paths are of the same color, pruned paths are showed as dashed line segments.
   (e): Extracted paths are shortened by Alg.\ref{alg:search}.
   (d): The shortened paths guide PGOs in parallel to generate several locally optimal trajectories. 
   Note that the red trajectory is the smoothest since it takes less turns, while the associated guiding path is longer than others.
   }
   \vspace{-0.6cm}
\end{figure}

Alg.\ref{alg:topo_roadmap} is used to construct a UVD roadmap $ \mathcal{G} $ capturing an abundant set of paths from different UVD classes.
Unlike standard PRM containing many redundant loops, our method generates a more compact roadmap where each UVD class contains just one or a few paths (displayed in Fig.\ref{fig:roadmap1}-\ref{fig:roadmap3}).

We introduce two different kinds of graph nodes, namely \textit{guard} and \textit{connector}, similar to the Visibility-PRM \cite{simeon2000visibility}.
The guards are responsible for exploring different part of the free space, and any two guards $ g_1 $ and $ g_2 $ are not \textit{visible} to each other (line $ \overline{g_1g_2} $ is in collision).
Before the main loop, two guards are created at the start point $ s $ and end point $ g $.
Every time a sampled point is invisible to all other guards, a new guard is created at this point (Line 6-7).
To form paths of the roadmap, connectors are used to connect different guards (Line 7-19).
When a sampled point is visible to exactly two guards, a new connector is created, either to connect the guards to form a topologically distinct connection (Line 19-20), or to replace an existing connector to make a shorter path (Line 16-17).
Limits of time ($t_{max}$) or sampling number ($ N_{max} $) are set to terminate the loop.

With the UVD roadmap, a depth-first search augmented by a visited node list is applied to search for the paths between $s$ and $g$, similar to \cite{rosmann2017integrated}.


\subsection{Path Shortening \& Pruning}
\label{subs:short_prune}
\begin{figure}[t]
	\begin{center}          
		\subfigure
		{\includegraphics[width=0.49\columnwidth]{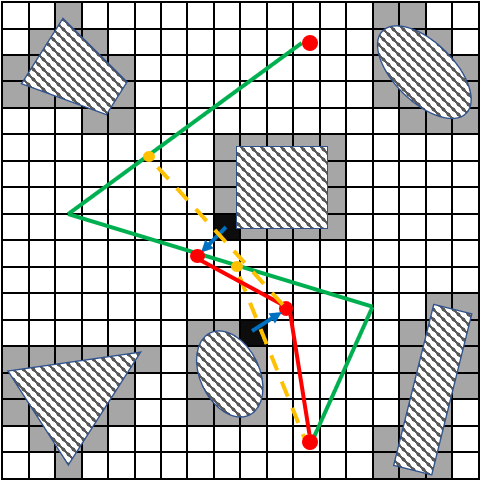}}       
		\subfigure
      {\includegraphics[width=0.49\columnwidth]{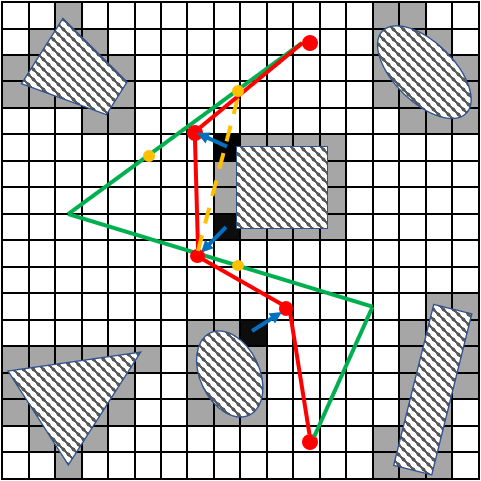}}       
      \vspace{-0.8cm}
	\end{center}
   \caption{\label{fig:shorten} A detoured and long path (green line) is shortened. 
   The yellow discretized point are not visible to the last waypoint of the shortened path (red). 
   The center points of the associated blocking voxel (black) are pushed away and appended as new waypoints.}
   \vspace{-0.1cm}
\end{figure} 
As shown in Fig. \ref{fig:roadmap4}, some paths obtained from Alg. \ref{alg:topo_roadmap} may be detoured. 
Such paths are unfavorable, since in the first phase of PGO it can deform a trajectory excessively and make it unsmooth. 
Hence, Alg. 2 find a topologically equivalent shortcut path $ \textit{P}_s $ for each $ \textit{P}_r $ found by the depth-first search (illustrated in Fig. \ref{fig:shorten}).
The algorithm uniformly \textbf{Discretize}s $ \textit{P}_r $ to a set of points $ \textit{P}_d $.
In each iteration, if a point $p_d$ in $ \textit{P}_d $ is invisible from the last point in $ \textit{P}_s $ (Line 3, 4), the center of the first occupied voxel blocking the view of $ \textit{P}_s.\textbf{back}() $ is found (Line 5).  
This point is then pushed away from obstacles in the direction orthogonal to $ l_d $ and coplanar to both $l_d$ and the ESDF gradient at $ p_b $ (Line 6), after which it is appended to $ \textit{P}_s $ (Line 7).
The process continues until the last point is reached.

Although in Alg. \ref{alg:topo_roadmap}, redundant connection between two guards are avoided, there may exist a small number of redundant paths between $s$ and $g$ (Fig. \ref{fig:roadmap4}).
To completely exclude repeated ones, we check the equivalence between any two paths and only preserve topologically distinct ones.
Also note that the number of distinctive paths grows exponentially with the number of obstacles.
In case of complex environments, it is computationally intractable to use all paths to guide parallel optimization.
For this reason, we only select the first $ K_{max} $ shortest paths.
Paths more than $ r_{max} $ times longer than the shortest one are also pruned away.
Such strategies bound the complexity and will not miss the potentially optimal solution, because a very long path is very unlikely to result in the optimal trajectory.

\begin{algorithm}[t]
   \DontPrintSemicolon
   $ \textit{P}_d \gets \tnb{Discretize}(\textit{P}_r), \ \textit{P}_s \gets \{\textit{P}_d.\tnb{front}() \} $\;
   \For{$ \tnb{each} \ p_d \in \textit{P}_d $}{
      $ l_d \gets \tnb{Line}(\textit{P}_s.\tnb{back}(), p_d) $\;
      \If{$ \lnot \ \tnb{LineVisib}(l_d) $}{
         $ p_b \gets \tnb{BlockPoint}(l_d) $\;
         $ p_o \gets \tnb{PushAwayObs}(p_b, l_d) $\;
         $ \textit{P}_s.\tnb{push\_back}(p_o) $\;
         }
   }
   $ \textit{P}_s.\textnormal{\textbf{push\_back}}(\textit{P}_d.\textnormal{\textbf{back}}()) $\;
\caption{Finding a topologically equivalent shortcut path $\textit{P}_s$ for $ \textit{P}_r $. \label{alg:search}}
\end{algorithm}

\section{Risk-aware Trajectory Refinement}
\label{sec:refinement}

Our trajectory refinement takes the best trajectory from the parallel PGO $ \mathbf{p}_i(t) $ as input, modifies it in its vicinity and outputs the refined trajectory $ \mathbf{p}_r(t) $ (detailed in Alg.\ref{alg:refine}).
It starts by checking the visibility status of $ \mathbf{p}_i(t) $, after which the visibility to relevant unknown space and the safe reaction distance are enforced in the iterative refinement.

\subsection{Checking Visibility Status}

\begin{figure}[t!]
	\begin{center}          
		{\includegraphics[width=0.7\columnwidth]{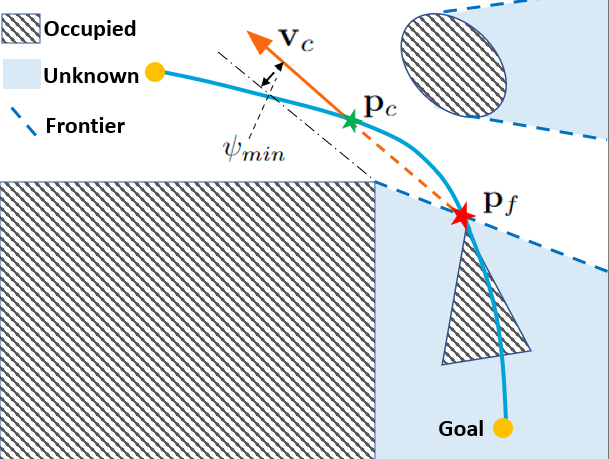}}       
	\end{center}
   \caption{\label{fig:visib} An illustration of checking the visibility status of the inputted trajectory.
   }
   \vspace{-1.5cm}
\end{figure}

The visibility status is encoded by several variables: $ t_f, \mathbf{p}_f $, $ t_c, \mathbf{p}_c, \mathbf{v}_c $, which are important information about the unknown space passed through by $\mathbf{p}_i(t)$.
Some of the involved variables are illustrated in Fig.\ref{fig:visib}. 

\subsubsection{Frontier Intersecting Point}

As the unknown environment is only partially observed, at some time $ t_f $ the trajectory $ \mathbf{p}_i(t) $ exits the known-free space and enters the unknown space.
The position $ \mathbf{p}_f = \mathbf{p}_i(t_f) $, should be prioritized for observation due to three reasons: 
First, it is highly relevant for the future flight, because it belongs to a promising trajectory going toward the goal.  
Second, it may be dangerous to the flight. In the worst case an unknown obstacle can be right adjacent to it. 
Third, it will be reached earlier compared to other unknown position along $\mathbf{p}_i(t)$.  
Therefore in $\tnb{FrontierIntersection}()$ we search along $\mathbf{p}_i(t)$ with a discrete time step, recording the first unknown point and the corresponding time.

\subsubsection{Visibility Metric}
\label{sssec:visibmetric}

During the flight, it is preferred that $ \mathbf{p}_f $ becomes visible to some preceding positions on $\mathbf{p}_i(t)$.
Quantitatively, we want some visibility level $ \psi $ of $ \mathbf{p}_f $ to be not less than an expected level $ \psi_{min} $, so that not only $ \mathbf{p}_f $ is visible but also the visibility level is tolerant to external disturbance. 
To gauge how reliable $ p_f $ is visible, we adopt the metric used by \cite{jeon2019online,tsai2004visibility}, defining the visibility level from a position $ \mathbf{p} \in \mathbb{R}^{3} $ to $ \mathbf{p}_f $ as:
\begin{equation}
\label{equ:visib}
   \psi(\mathbf{p},\mathbf{p}_f) = \min_{\mathbf{q} \in l(\mathbf{p}, \mathbf{p}_f)} d(\mathbf{q})   
\end{equation}
which represents the the smallest Euclidean signed distance between the line segment $l(\mathbf{p}, \mathbf{p}_f)$ and obstacles.
The evaluation of Equ.\ref{equ:visib} requires traversing $l(\mathbf{p}, \mathbf{p}_f)$ and checking the Euclidean signed distance for each point.
Fortunately, an ESDF derived from the occupancy grid map is maintained by our mapping module for supporting the trajectory optimization (Sect.\ref{sec:pgo}), so minimal distance can be queried in constant time.

\subsubsection{Critical View Direction}
\label{sssec:critical}

We are interested in whether $ \mathbf{p}_f $ can be viewed from preceding positions on $ \mathbf{p}_i(t) $, i.e., if $ \psi(\mathbf{p}_i(t_k),\mathbf{p}_f) \ge \psi_{min} $ for some time $ t_k \le t_f $.
The best case is that for all $ t \le t_f $ the visibility level is higher than $ \psi_{min} $, in which no modification to the trajectory is needed (Line 3-4).
However, in cluttered environments $ \mathbf{p}_f $ may not be observable until some time $ t_c $, which means $ \psi(\mathbf{p}_i(t),\mathbf{p}_f) < \psi_{min}$ when $ t < t_c $ and $ \psi(\mathbf{p}_i(t),\mathbf{p}_f) \ge \psi_{min}$ when $ t \ge t_c $.
Here, the view direction $ \mathbf{v}_c = \frac{\mathbf{p}_c - \mathbf{p}_f}{\Vert \mathbf{p}_c - \mathbf{p}_f \Vert} $ where $ \mathbf{p}_c = \mathbf{p}_i(t_c) $ is a critical view direction, in which $ \psi(\mathbf{p}_i(t),\mathbf{p}_f) $ just reaches the desired level $ \psi_{min} $.
In this case $\tnb{CriticalView}()$ reports $\mathbf{v}_c$ and the corresponding position $\mathbf{p}_c$ and time $t_c$.



\begin{figure}[t!]
	\begin{center}          
		\subfigure
		{\includegraphics[width=0.49\columnwidth]{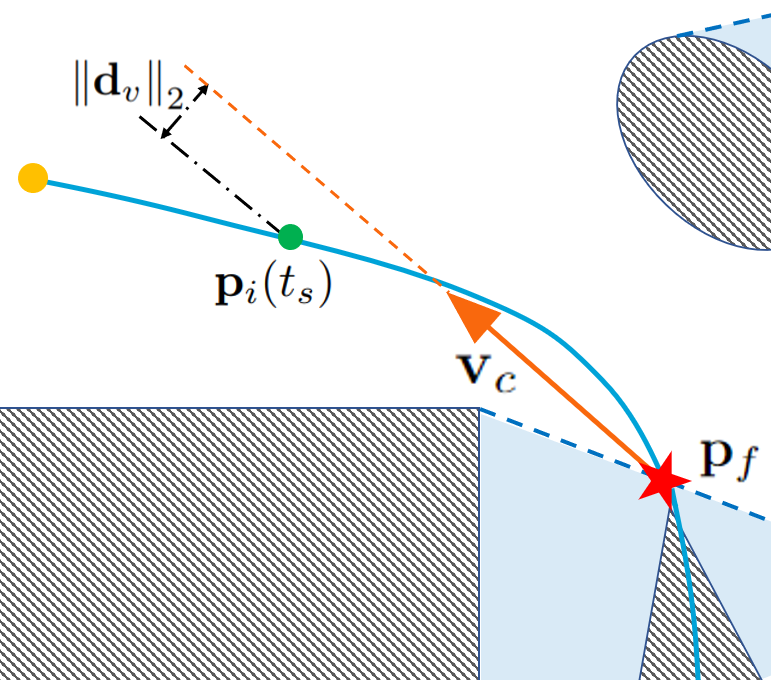}}       
		\subfigure
      {\includegraphics[width=0.49\columnwidth]{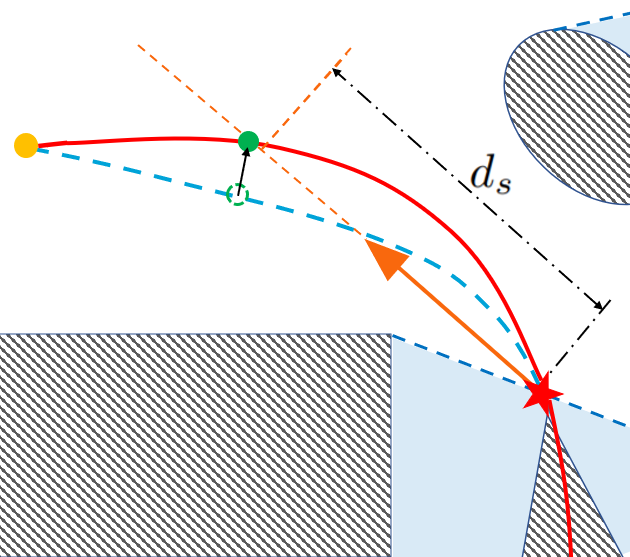}}       
	\end{center}
   \caption{\label{fig:refine} The initial trajectory (blue) gets refined with the visibility status.
   Along the refined trajectory (red) $ \mathbf{p}_f $ will become viewable earlier, wherein the reaction distance for collision avoidance is also increased.
   }
\end{figure} 

\begin{algorithm}[t]
   \DontPrintSemicolon
   $\mathbf{p}_f, t_f \gets \tnb{FrontierIntersection}(\mathbf{p}_i(t)) $\;
   $visible, \mathbf{v}_c, \mathbf{p}_c, t_c \gets \tnb{CriticalView}(\mathbf{p}_i(t), \mathbf{p}_f) $\;
   \If{$ visible == True $}{
      \Return{$\mathbf{p}_i(t)$} \;
   }
   $v_c \gets \left\| \dot{\mathbf{p}}_i(t_c) \right\|_{2}, \ d_{cf} \gets \left\| \mathbf{p}_c - \mathbf{p}_f \right\|_{2}    $ \;
   \If{$ {v_{c}^{2}}/{2 a_{max}} \le d_{cf} - R_q $}{
      \Return{$\mathbf{p}_i(t)$} \;
   }
   $\hat{v}_s \gets \tnb{AverageSpeed}(\mathbf{p}_i(t), t_0, t_f)$ \;
   \Repeat{$ {v_{s}^{2}}/{2 a_{max}} \le d_{sf} - R_q $}{
      $\hat{d}_{s} \gets {\hat{v}_{s}^{2}}/{2 a_{max}} + R_q $ \;
      $t_s \gets \underset{t}{\arg \min} \ f_{\mathbf{d}_v} + w_r f_{\mathbf{d}_{sf}} $ \;
      $\mathbf{p}_r(t) \gets  \tnb{RefineTrajectory}(\mathbf{p}_i(t), \mathbf{p}_f, \mathbf{v}_c, t_s, \hat{d}_s)$ \;
      $v_s \gets \left\| \dot{\mathbf{p}}_r(t_s) \right\|_2, \ d_{sf} \gets \left\| \mathbf{p}_r(t_s) - \mathbf{p}_f \right\|_{2}$\;
      $\hat{v}_s \gets \alpha \cdot \hat{v}_s$ \;
   }
   \Return{$\mathbf{p}_r(t)$} \;
\caption{Risk-aware Trajectory Refinement. \label{alg:refine}}
\end{algorithm}

\subsection{Iterative Refinement}
\label{subs:optimize}

As is the worst case, an unknown obstacle may be revealed right behind $\mathbf{p}_f$ and block the trajectory.
To ensure safety, we refine $\mathbf{p}_i(t)$ so that under the worst case the quadrotor will be able to avoid collision by taking some maneuvers whose single-axis acceleration does not exceed the limit $ a_{max} $.
We first check whether $\mathbf{p}_i(t)$ satisfies the worst-case safety criteria (Line 5-7). 
If it does not, the critical view and safe reaction distance constraints are enforced (Line 8-15).


\subsubsection{Worst-case Safety Criteria}

As showed in Sect.\ref{sssec:critical}, initially $ \mathbf{p}_f $ will not become reliably viewable until $ t_c $ at $ \mathbf{p}_c $.
Suppose at $ \mathbf{p}_c $ the speed is $ v_c $, while the distance to $ \mathbf{p}_f $ is $ d_{cf} $. 
Similar to \cite{liu2016high}, we check if Equ.\ref{equ:criteria} holds: 
\begin{equation}
   \label{equ:criteria}
   {v_{c}^{2}}/{2 a_{max}} \le d_{cf} - R_q
\end{equation}
which means that if at $\mathbf{p}_c$ the quadrotor sees an obstacle, it can decelerate to a stop before colliding with the obstacle right behind $\mathbf{p}_f$. 
$ R_q $ compensates the quadrotor size and disturbance.
If it is not true, extra constraints are added to meet this criteria.

\subsubsection{View and Safety Constraints}
  
If initially it would be too late for collision avoidance, $ \mathbf{p}_f $ should be viewed at an earlier stage $ t_s < t_c $, so that maneuvers to avoid collision can be taken in advance.
Given the critical view direction $ \mathbf{v}_c $, this constraint can be expressed as:
\begin{align}
\label{equ:cons1}
   \left\|\mathbf{d}_{v}\right\|_{2} & \leq \delta_{v}  \\ 
   \mathbf{d}_{v} = \left(\mathbf{p}_{i}(t_s)-\mathbf{p}_{f}\right) &- \left(\left(\mathbf{p}_{i}(t_s)-\mathbf{p}_{f}\right) \cdot \mathbf{v}_{c}\right) \mathbf{v}_{c}
\end{align}
The interpretation is that at $ t_s $ the quadrotor should have already reached the ray $ \left\{ \mathbf{q} \, | \, \mathbf{q} = \mathbf{p}_f + \lambda \mathbf{v}_c, \lambda \ge 0 \right\} $ from which $ \mathbf{p}_f $ is viewable (Fig.\ref{fig:refine}).
Besides the advanced observation, the distance from $ \mathbf{p}_r(t_s) $ to $ \mathbf{p}_f $ should be sufficient for an emergency stop:
\begin{equation}
\label{equ:cons2}
   {d}_{sf} = \left\| \mathbf{p}_{i}(t_s)-\mathbf{p}_{f}\right\|_2 \ge d_s
\end{equation}
where $ d_s $ is the safe reaction distance that depends on the speed $ v_s $ at the refined trajectory $ \mathbf{p}_r(t_s) $: $ d_{s}= {v_{s}^{2}}/{2 a_{max}} + R_q $.
However, since the refined trajectory is not obtained yet, $ v_s $ is not available at this stage. 
To solve this chicken-and-egg problem, we introduce an iterative strategy.
At the beginning we use the average speed of the segment of $ \mathbf{p}_i(t) $ between $ \left[ t_0, t_f \right] $ as an estimate of $ v_s $ (Line 8), where $t_0$ is the start time of $ \mathbf{p}_i(t) $.
Then the trajectory is refined with the view and safety constraints (Line 10-12).
After the refinement we check whether the new trajectory satisfies the safety criteria. 
If it does not, we increased the estimated speed $\hat{v}_s$ with a factor $\alpha$ slightly larger than 1 and redo the refinement (Line 13-15).
This strategy is complete because it only terminates after the safety criteria is indeed met.
It also terminates quickly, since the speed along the smooth trajectory varies slightly and it only takes one or a few steps to find a good estimate of $v_s$ in practice. 

$\tnb{RefineTrajectory}()$ is essentially optimizing the trajectory with the newly introduced view and safety constraints.
To incorporate them into our gradient-based optimization (Sect.\ref{subs:pro_form}), Equ.\ref{equ:cons1}-\ref{equ:cons2} are soft-constrained by penalty functions:
\begin{align}
   f_{\mathbf{d}_v} &= \mathcal{F}(\delta_v, \left\| \mathbf{d}_{v} \right\|_2) \\ 
   f_{\mathbf{d}_{sf}} &= \mathcal{F}(\left\| \mathbf{p}_{i}(t_s)-\mathbf{p}_{f}\right\|_2, d_s)
\end{align}
Here $ \mathcal{F}() $ is the penalty function (Equ.\ref{equ:potential}).
The Jacobian of these terms are:
\begin{equation}
   \frac{\partial f_{\mathbf{d}_v}}{\partial \mathbf{q}_{k}} = \left\{
      \begin{aligned}
          \frac{2 (\left\| \mathbf{d}_{v} \right\|_2-\delta_v) \mathbf{d}_{v}^{\mathrm{T}}}{\left\| \mathbf{d}_{v} \right\|_2} (\mathbf{I} - \mathbf{v}_c \mathbf{v}_{c}^{\mathrm{T}}) \, \frac{\partial \mathbf{p}_{i}(t_s)}{\partial \mathbf{q}_{k}}, & \left\| \mathbf{d}_{v} \right\|_2 > \delta_v \\
          0 \quad \quad \qquad \qquad \qquad, \, & \text{else}
      \end{aligned}
      \right.
\end{equation}
\begin{equation}
   \frac{\partial f_{\mathbf{d}_{sf}}}{\partial \mathbf{q}_{k}} = \left\{
      \begin{aligned}
         & \qquad \qquad \qquad \, \ \ 0 \qquad \qquad, \left\| \mathbf{p}_{i}(t_s)-\mathbf{p}_{f}\right\|_2 \ge d_s \\
         & \frac{2(\left\| \mathbf{p}_{i}(t_s)-\mathbf{p}_{f}\right\|_2 - d_s) \mathbf{p}_i(t_s)^{\mathrm{T}}}{\left\| \mathbf{p}_{i}(t_s)-\mathbf{p}_{f}\right\|_2} \frac{\partial \mathbf{p}_{i}(t_s)}{\partial \mathbf{q}_{k}} , \, \text{else}
      \end{aligned}
      \right.
\end{equation}
Here $ \mathbf{q}_{k} $ B-spline control points associated with $ \mathbf{p}_i(t_s) $. 
The refinement minimizes the cost function:
\begin{align}
   f_{r}= \underbrace{\lambda_{s} f_{s} +\lambda_{c} f_{c}+ \lambda_{d}\left( f_{v} + f_{a} \right)}_{f_{p2}} + \lambda_{r} (f_{\mathbf{d}_v} + w_{r} f_{\mathbf{d}_{sf}}) 
\end{align}
where $f_{p2}$ is exactly the same as Equ.\ref{equ:fp2}.
Note that applying the constraints to different $t_s$ generates different results.
However, including $t_s$ into the optimization to find its best value would make the problem too difficult to be solved quickly.
For simplicity, we first determine $ t_s $ as the time that minimizes $ f_{\mathbf{d}_v} + w_{r} f_{\mathbf{d}_{sf}} $ for the initial $ \mathbf{p}_i(t) $ and use it for the refinement (Line 11-12). 
Quantitatively, the selected $t_s$ leads to minimal violation of the constraints, therefore enforcing the constraints at $t_s$ also requires less modification to $ \mathbf{p}_i(t) $, which is a reasonably good choice.

\section{Yaw Angle Planning}
\label{sec:yaw}

Quadrotors typically have limited sensor FOVs.
To improve flight safety, we plan the trajectory of yaw angle to actively observe the environments.
Inspired by the recent two-step quadrotor motion planning paradigm \cite{CheSuShe2015, fei2018icra, fei2016ssrr, fei2018jfr, ding2019efficient,zhou2019robust}, we also decompose the yaw angle planning into a graph search problem and a trajectory optimization.

\begin{figure}[t!]
	\begin{center}          
		{\includegraphics[width=0.9\columnwidth]{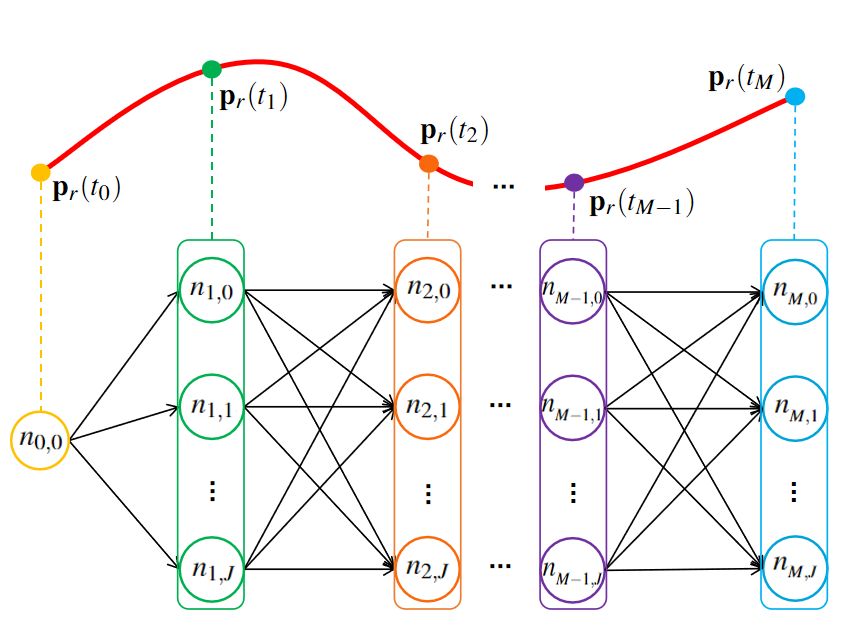}}       
	\end{center}
   \caption{\label{fig:search} The directed graph modeled for searching a sequence of yaw angles along the refined trajectory. Each layer of nodes (with the same color) is associated with a position on the trajectory.  
   }
\end{figure} 

\subsection{Graph Search}

\subsubsection{Problem Modeling}
We model a graph search problem to seek for an sequence of yaw angles $ {\Xi} := \left\{ \xi_0, \xi_1, \cdots, \xi_M \right\} $ along the refined trajectory that trades off the smoothness and information gain (IG, detailed in Sect.\ref{sssec:gain}) of the unknown space.
Given the quadrotor trajectory $ \mathbf{p}_r(t) $ (Sect.\ref{sec:refinement}), a set of positions $ \mathbf{p}_{r, i}, i \in \left[ 0,1,\cdots, M \right] $ uniformly distributed along the trajectory at $ \left\{ t_0, t_1, \cdots, t_M \right\} $ are selected.
At $ \mathbf{p}_i $ expect $ i = 0 $, where the yaw angle is already determined by the current quadrotor's state, several graph nodes $ {n}_{i, j}, j \in \left[0, 1, \cdots, J \right] $ are created, each of which associates a different angle $ \xi_{i, j} $ and the IG $ g_{i, j} $ at the state $ \left( \mathbf{p}_{r, i}, \xi_{i, j} \right) $.
For each pair of nodes $ n_{i,j_1}, n_{i+1, j_2} $ associated with adjacent positions, a graph edge from $ n_{i,j_1} $ to $ n_{i+1, j_2} $ is created.
This process construct a directed graph as shown in Fig.\ref{fig:search}. 
To find a sequence of yaw that maximizes IG and smoothness, we search a path in the graph that minimizes the cost $c(\Xi)$, which can be solved efficiently by the Dijkstra algorithm:
\begin{equation}
\label{equ:pathcost}   
   c(\Xi) = \sum_{i=1}^{M} \left\{ -g_{i, j_i} + \mu \,  (\xi_{i,j_i} - \xi_{i-1,j_{i-1}})^2 \right\}
\end{equation}
where $ \mu $ is used to adjust the weighting of smoothness.

\subsubsection{Information Gain}
\label{sssec:gain}

\begin{figure}[t!]
   \begin{center}          
      \subfigure[\label{fig:gain}]
		{\includegraphics[width=0.49\columnwidth]{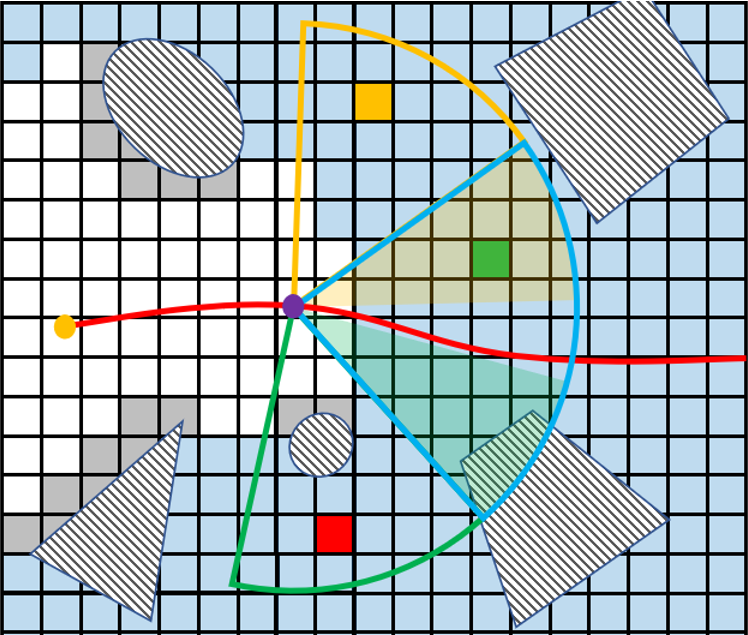}}       
      \subfigure[\label{fig:bias}]
		{\includegraphics[width=0.49\columnwidth]{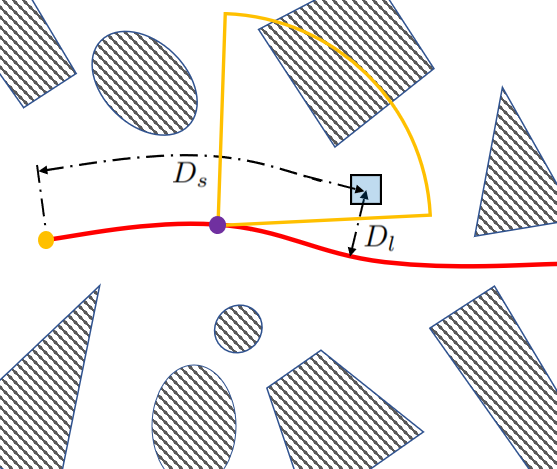}}       
	\end{center}
   \caption{\label{fig:gain12} 
   (a) The evaluation of potential information gain.
   Blue transparent voxels represent unknown areas. Occupied and free space is represented by gray and white colors.
   The yellow voxel will contribute to the information gain, while the red voxel will not since it is blocked by occupied voxels. 
   The green voxel is inside the overlapping region of multiple FOVs, whose visibility is stored after the first raycasting and queried directly in subsequent check. 
   (b) Weighting a unknown and visible voxel by its lateral and longitudinal distance to the trajectory.
   }
   \vspace{-0.5cm}
\end{figure}

We employ a similar method to \cite{bircher2018receding} which assesses potential IG as the number of unmapped voxels that comply with the camera model and are visible (not blocked by occupied voxels). 
However, the original method does raycasting for every voxels inside the camera FOV to validate their visibility, which is too expensive to function online.
Therefore, we adapt it to better suit the real-time planning in several ways:
(a) As is in \cite{oleynikova2018safe}, voxels inside the FOV are subsampled to approximate the actual gain, which leads to only slight error but great run time reduction.
(b) The gains of different $ \xi_{i,j} $ are evaluated in parallel.
(c) We borrow the techniques from \cite{millane2018c} to avoid repeated raycasting. 
As depicted in Fig.\ref{fig:gain}, we notice that at one position $ \mathbf{p}_{r, i} $ where different $ \xi_{i,j} $ are assessed, many voxels are in overlapping areas and are checked for visibility more than once.
To avoid unnecessary repetition, we store the visibility of each voxel when it is checked for the first time, so that in subsequent check the visibility and be queried directly.
In these ways, the overall IG evaluation time is reduced by over two orders of magnitude.           

In the context of exploration\cite{bircher2018receding}, every voxel contributes equally to the IG of one quadrotor configuration, ensuring that all space can be covered by the sensors uniformly. 
However, in a point-to-point navigation we do not aim at full coverage but prefer focusing on space relevant to the flight.
In particular, unknown voxels closer to the trajectory and the current position have higher influence to the flight.
Therefore, we use Equ.\ref{equ:bias} to bias the IG contribution $ Ig $ of a visible voxel centered at $ \mathbf{m} $:
\begin{equation}
\label{equ:bias}   
   Ig(\mathbf{m}) = \exp \left\{-w_l \, D_l(\mathbf{m}, \mathbf{p}_r(t)) -w_s \, D_s(\mathbf{m}, \mathbf{p}_r(t)) \right\}
\end{equation}
where $ D_l(\mathbf{m}, \mathbf{p}_r(t)) $ and $ D_s(\mathbf{m}, \mathbf{p}_r(t)) $ represent the lateral and longitudinal distance to trajectory $ \mathbf{p}_r(t) $ (Fig.\ref{fig:bias}).


\subsection{Yaw Angle Optimization}

Given the optimal path $ \Xi $ searched through the graph, we compute the trajectory of yaw angle $ \phi(t) $ that is smooth, dynamically feasible and passes through the sequential angles $ \xi_{j} $. 
We parameterize $ \phi(t) $ as a uniform B-spline with control points $ \Phi := \left\{ \phi_{c,0}, \phi_{c,1}, \cdots, \phi_{c,N_c} \right\} $ and knot span $ \delta t_{\phi} $. 
In this way, the convex hull property can be employed to ensure dynamic feasibility\cite{zhou2019robust}. 
The problem is formulated as:
\begin{align}
   & \underset{\Phi}{\arg \min } \int_{t_{0}}^{t_{M}} (\dddot{\phi}(t))^{2} d t + \gamma_{1} \sum_{i=0}^{M} ( \phi(t_i) - \xi_i )^2 \\ \nonumber
   & + \gamma_2 \left\{ \sum_{j=0}^{N_c-1} \mathcal{F}(\dot{\phi}_{max},\vert \dot{\phi}_{c,j} \vert) + 
   \sum_{j=0}^{N_c-2} \mathcal{F}(\ddot{\phi}_{max}, \vert \ddot{\phi}_{c,j} \vert) \right\}
\end{align}
Here the first term represents smoothness and the second term is a soft waypoint constraint enforcing $ \phi(t) $ to pass through $ \Xi $.
The last two terms are the soft constraints for dynamic feasibility, wherein $ \dot{\phi}_{c,j} $ and $ \ddot{\phi}_{c,j} $ are the B-spline control points of angular velocity and acceleration:
\begin{equation}
   \dot{\phi}_{c,j} = \frac{\phi_{c,j+1} - \phi_{c,j}}{\delta t_{\phi}}, \ \ 
   \ddot{\phi}_{c,j} = \frac{\phi_{c,j+2} - 2\phi_{c,j+1} + \phi_{c,j}}{\delta t_{\phi}^{2}}
\end{equation}
Thanks to the convex hull property of B-spline, the entire trajectory is guaranteed to be feasible given that the control points do not exceed the dynamic limits $ \dot{\phi}_{max}, \ddot{\phi}_{max} $.

\begin{figure}[t]
	\begin{center}          
		\subfigure[\label{fig:active_scene_1}]
		{\includegraphics[width=0.45\columnwidth]{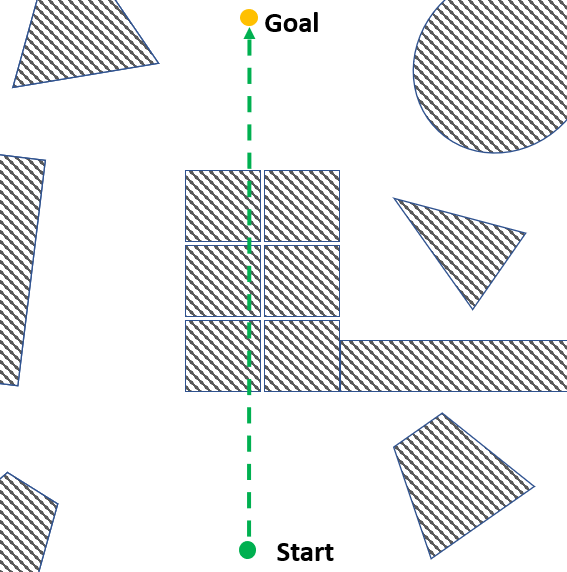}}       
		\subfigure[\label{fig:active_scene_2}]
      {\includegraphics[width=0.45\columnwidth]{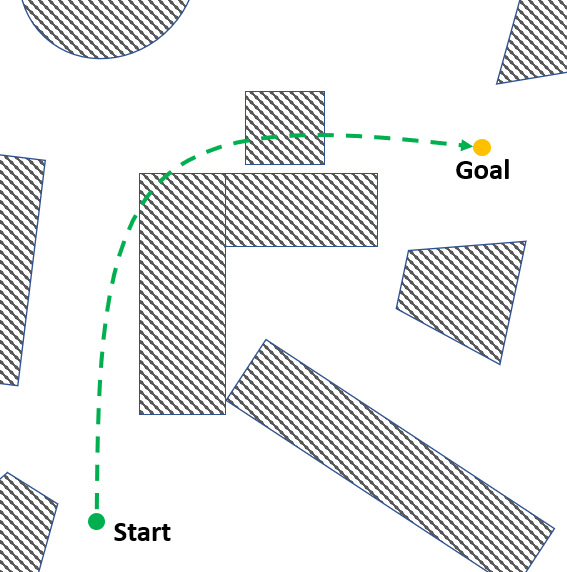}}       
	\end{center}
   \caption{\label{fig:active_scene} An illustration of the two real-world scenarios for testing the perception-aware replanning.
   (a) A large obstacles is placed along the straight-line global reference trajectory.
   (b) An obstacles is placed right behind the corner.
   }
   \vspace{-3.5cm}
\end{figure}

\begin{figure*}[htb]
	\begin{center}          
		\subfigure[\label{fig:4scene_1}]
		{\includegraphics[width=0.515\columnwidth]{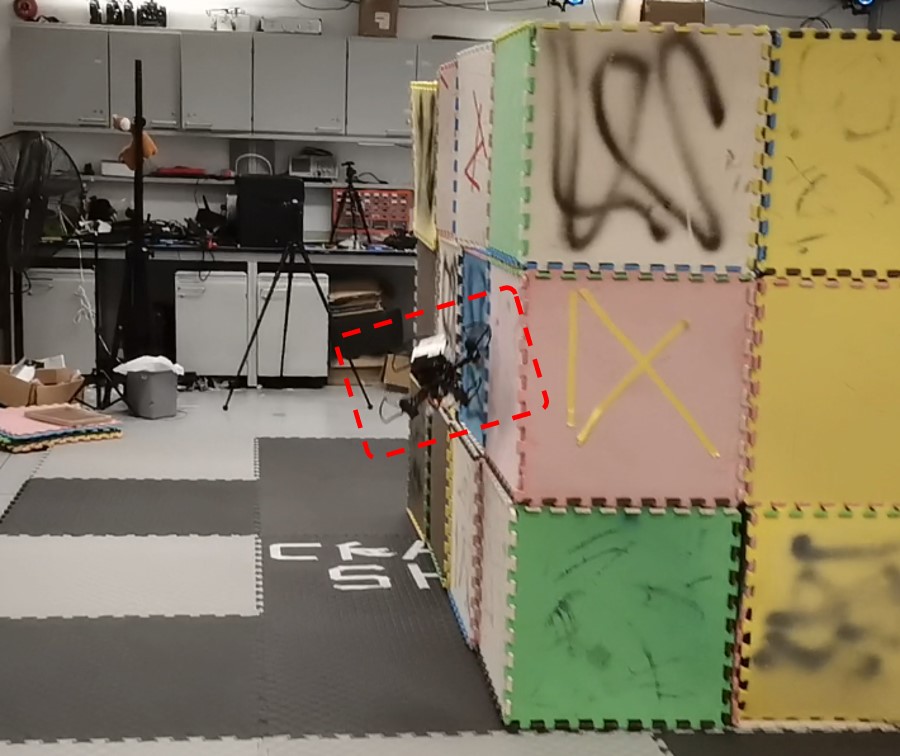}}       
		\subfigure[\label{fig:4scene_2}]
		{\includegraphics[width=0.465\columnwidth]{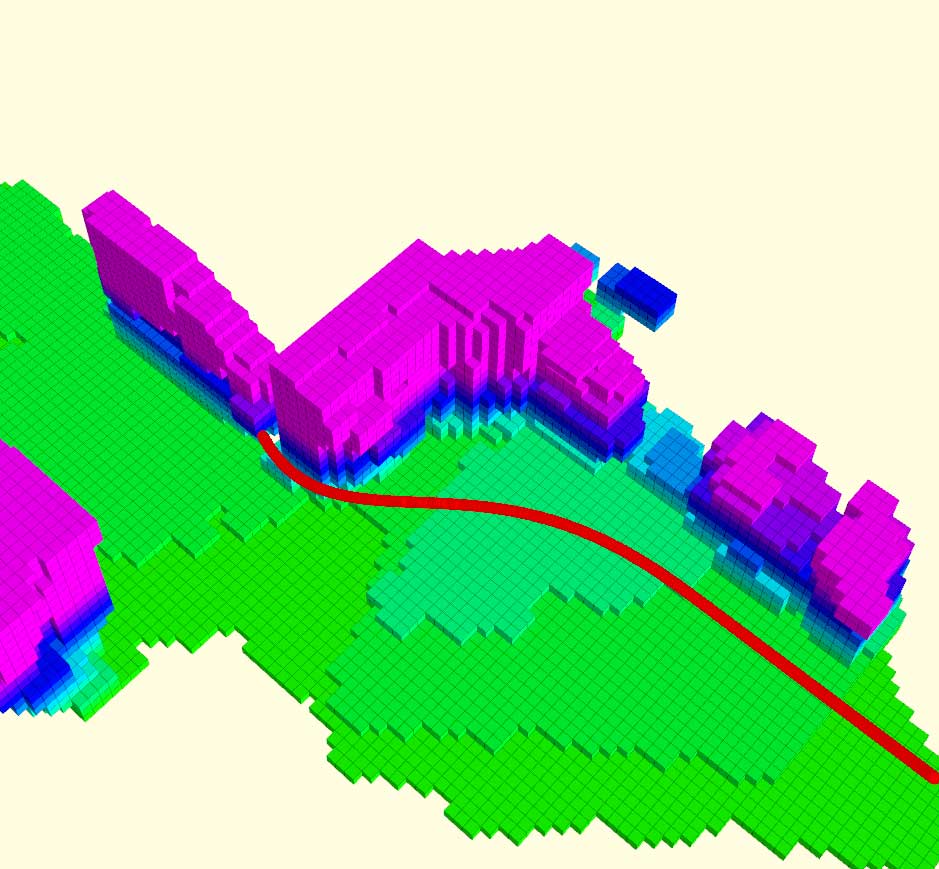}}       
		\subfigure[\label{fig:4scene_3}]
		{\includegraphics[width=0.515\columnwidth]{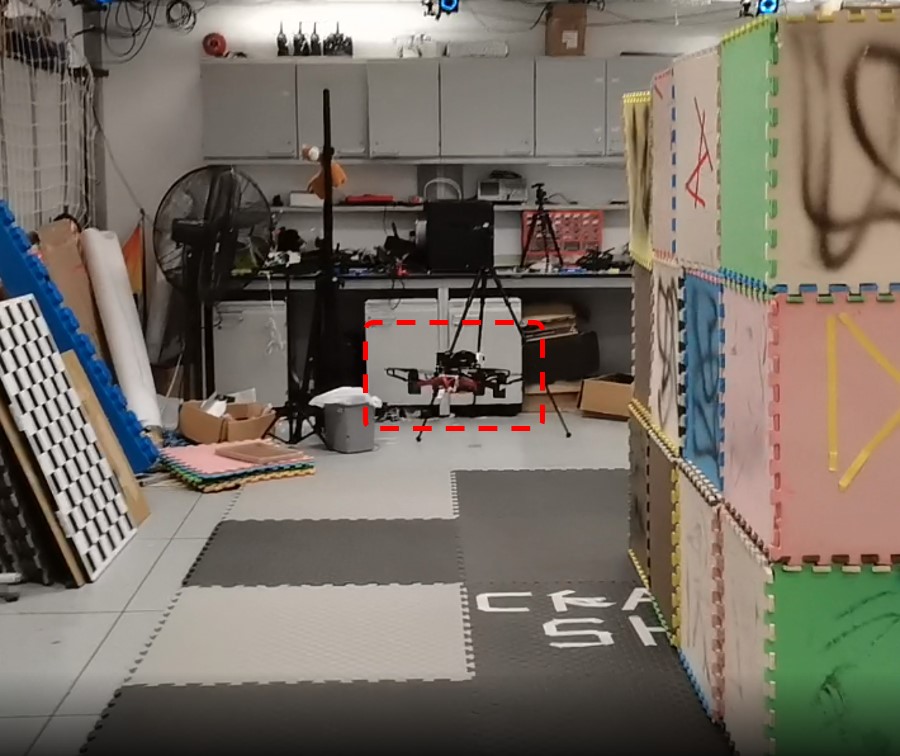}}       
		\subfigure[\label{fig:4scene_4}]
		{\includegraphics[width=0.465\columnwidth]{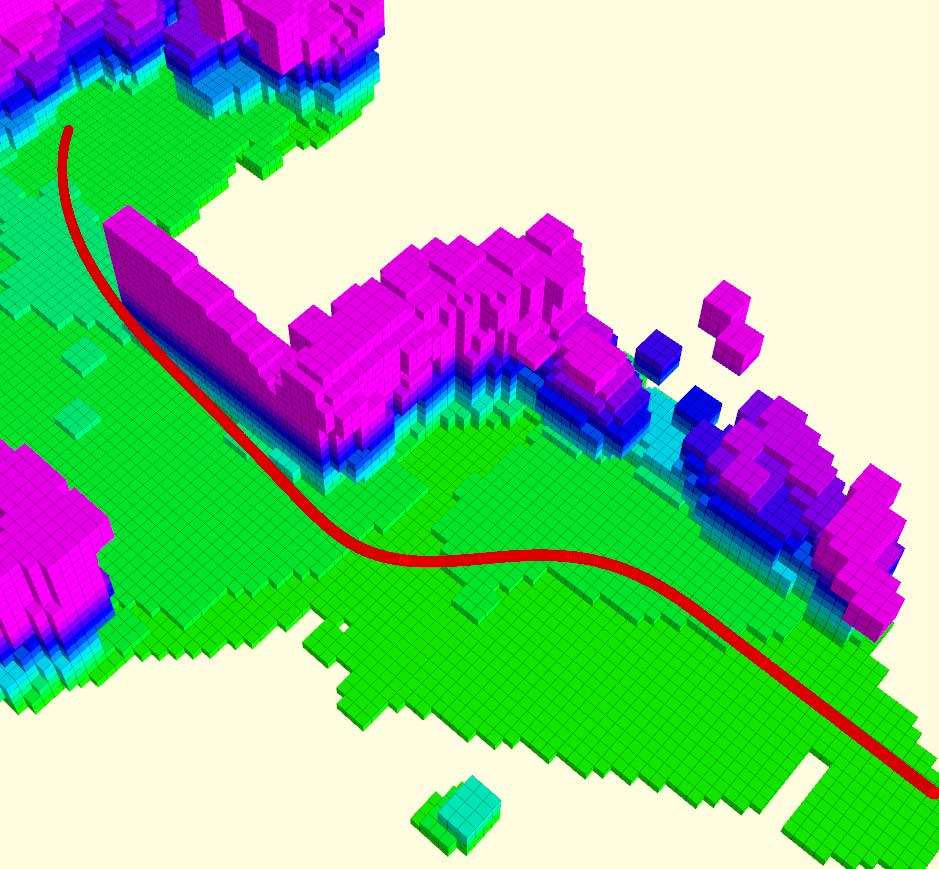}}       
      \vspace{-0.5cm}
	\end{center}
   \caption{\label{fig:4scene} Comparison of four local planners in scene 1.
   (a) using planners A and B, the boxes in the back row are not discovered until the quadrotor gets very close. 
   Consequently the quadrotor has to pause in emergency.
   (b) The 3D maps (colorful voxels) and executed trajectories (red curves) at the moment when \textbf{boxes in the back are just discovered}.
   (c),(d): the quadrotor avoid all boxes successfully with planners D.
   More details are showed in the video.
   }
   \vspace{-0.1cm}
\end{figure*} 

\begin{figure}[htb]
	\begin{center}          
		\subfigure[\label{fig:comp_refine_wo}]
		{\includegraphics[width=0.495\columnwidth]{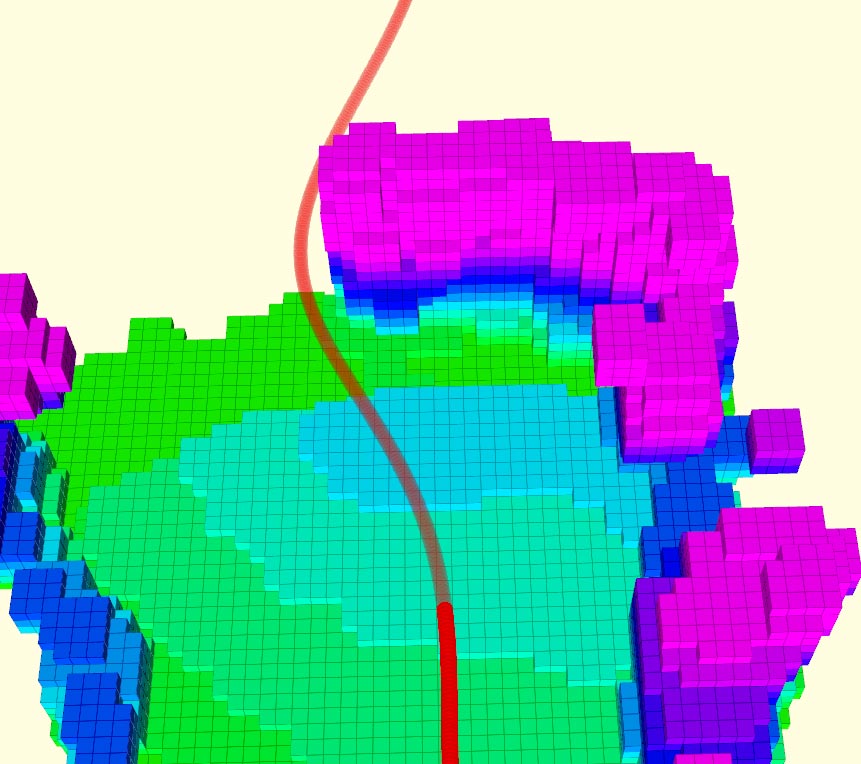}}       
		\subfigure[\label{fig:comp_refine_w}]
      {\includegraphics[width=0.485\columnwidth]{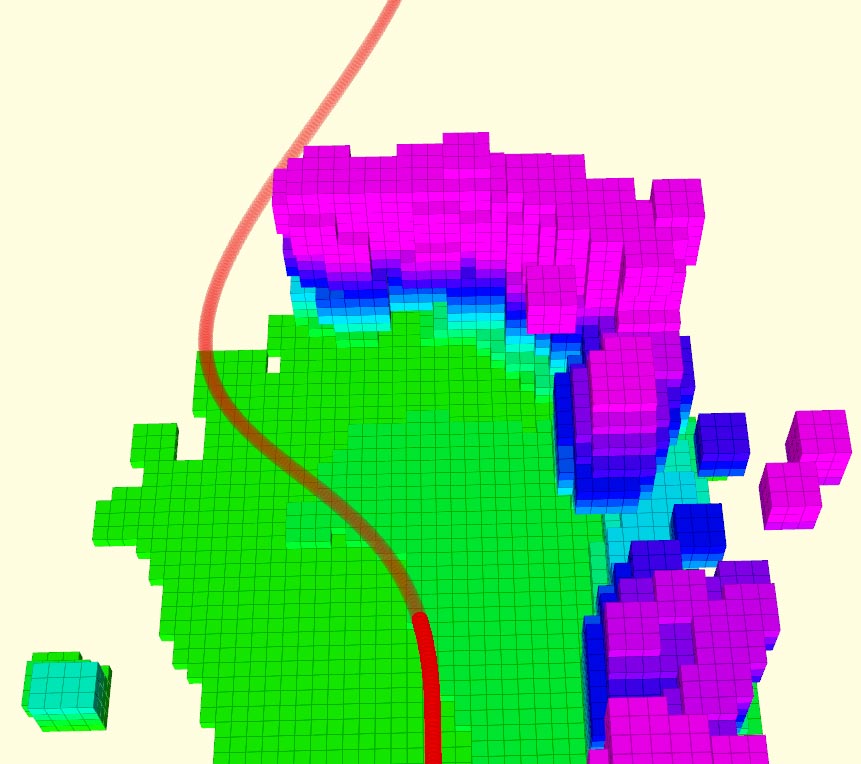}}       
      \vspace{-0.8cm}
	\end{center}
   \caption{\label{fig:comp_refine} A comparison of the trajectories in scene 1 replanned with (a) optimistic assumption and (b) risk-aware refinement.
   Trajectories already executed and not executed yet are showed in opaque red and transparent red respectively.
   (a) Along the trajectory visibility toward the unknown region behind the observed obstacle is poor.
   (b) The trajectory deviates more laterally, therefore along it there is greater visibility toward the unknown area in the back. 
   }
   \vspace{-0.5cm}
\end{figure} 

\begin{figure}[htb]
	\begin{center}          
      {\includegraphics[width=0.49\columnwidth]{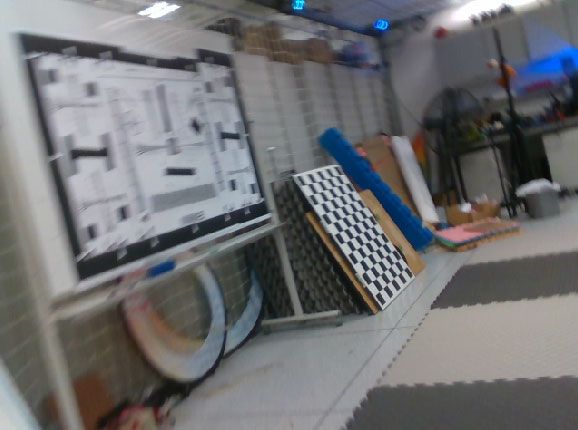}}       
      {\includegraphics[width=0.49\columnwidth]{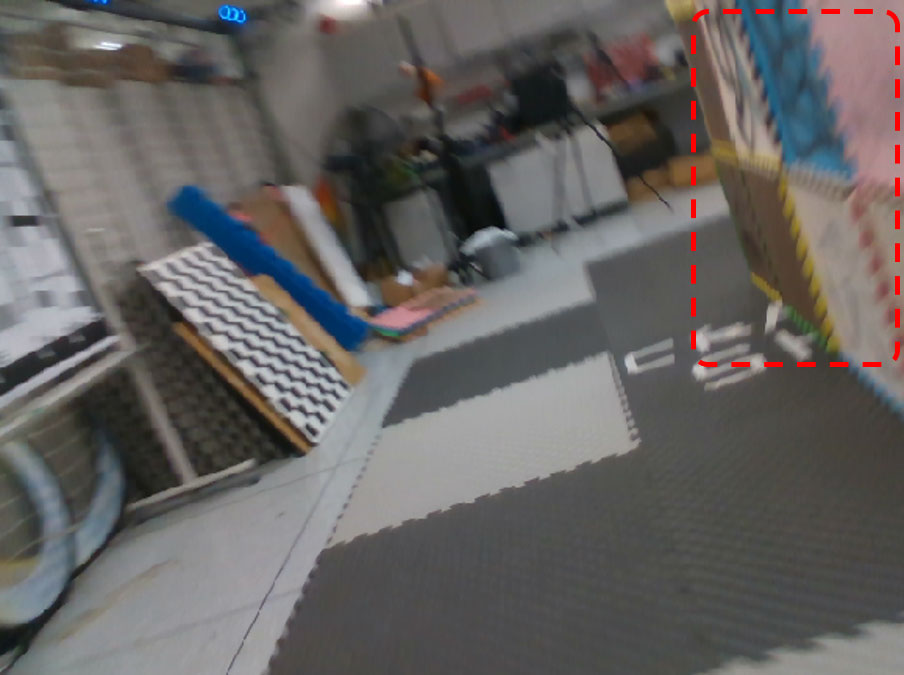}}       
		\subfigure[\label{fig:comp_yaw_1}]
		{\includegraphics[width=0.492\columnwidth]{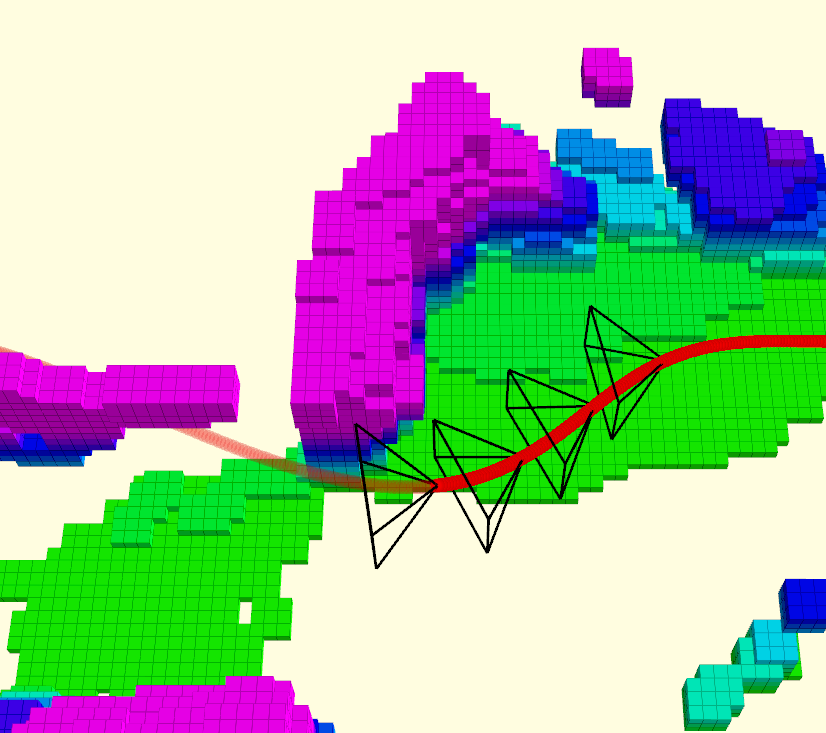}}       
		\subfigure[\label{fig:comp_yaw_2}]
      {\includegraphics[width=0.488\columnwidth]{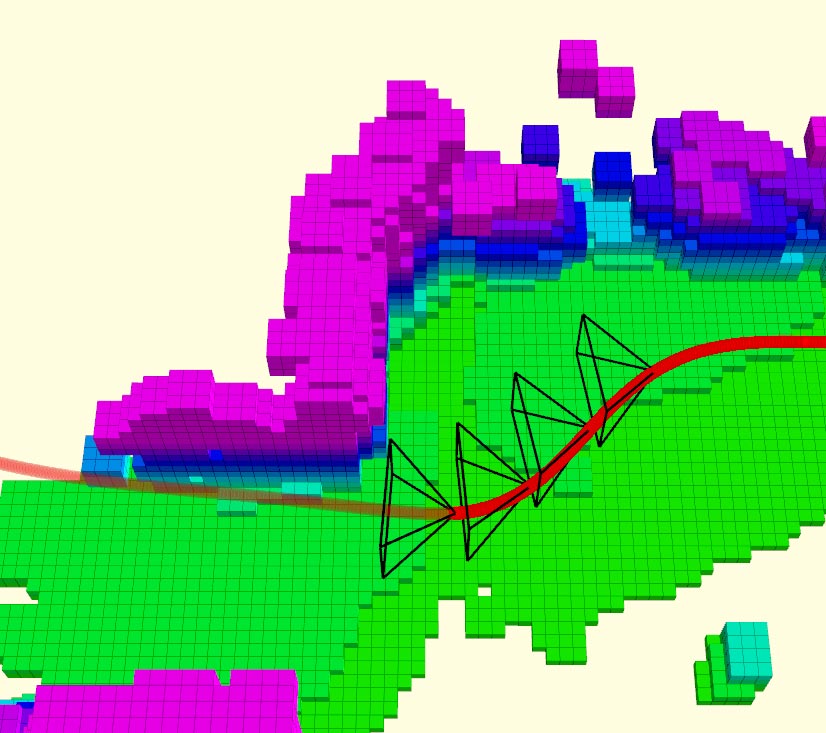}}       
      \vspace{-0.8cm}
	\end{center}
   \caption{\label{fig:comp_yaw} A comparison in scene 1 of (a) velocity-tracking yaw and (b) active exploration yaw. 
   Top: The first person view (FPV) images.  
   Bottom: the 3D maps and trajectories associated with the FPV images. 
   The camera poses within 0.9 seconds (interval $= 0.3 s$) are also visualized.  
   (a) The quadrotor face to the left side at the beginning, so unknown boxes on the right side are not mapped completely.
   (b) The quadrotor discovers the previously occluded boxes earlier. 
   }
   \vspace{-0.1cm}
\end{figure} 

\begin{figure*}[t]
	\begin{center}          
		\subfigure[\label{fig:scene2_1}]
      {\includegraphics[width=0.485\columnwidth]{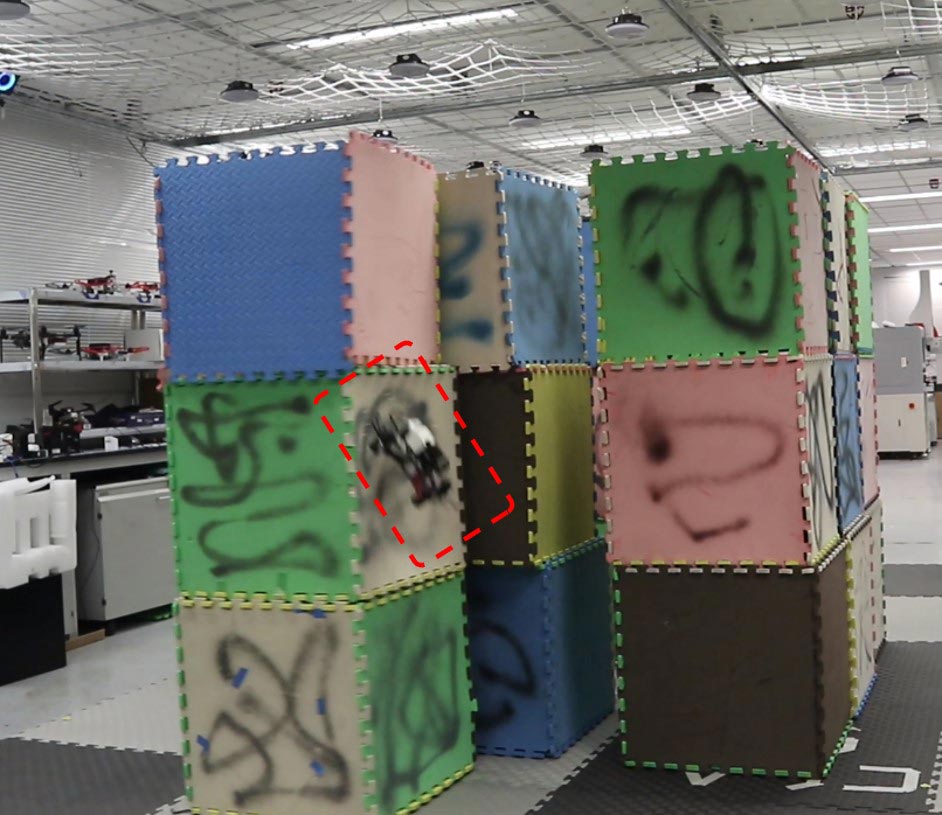}}       
		\subfigure[\label{fig:scene2_2}]
		{\includegraphics[width=0.495\columnwidth]{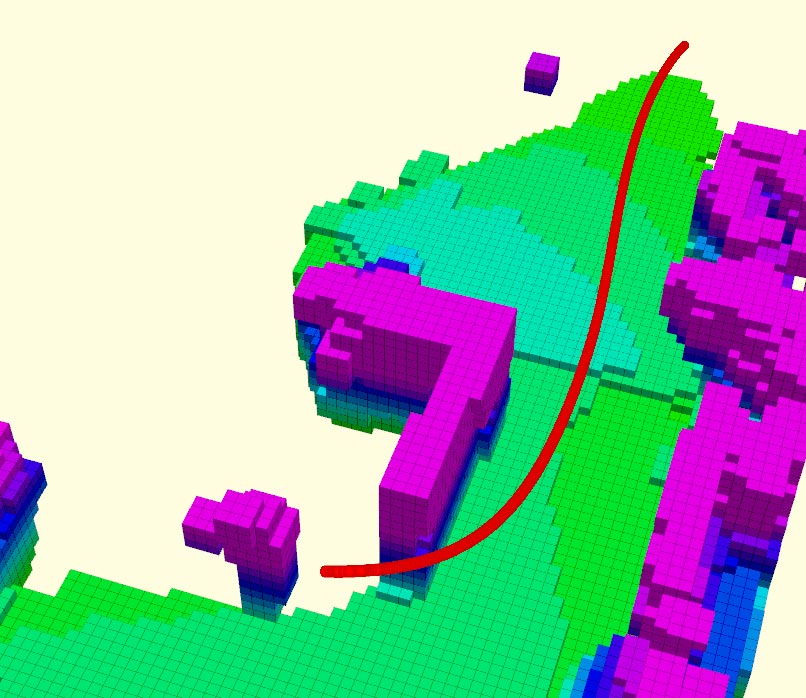}}       
		\subfigure[\label{fig:scene2_3}]
      {\includegraphics[width=0.485\columnwidth]{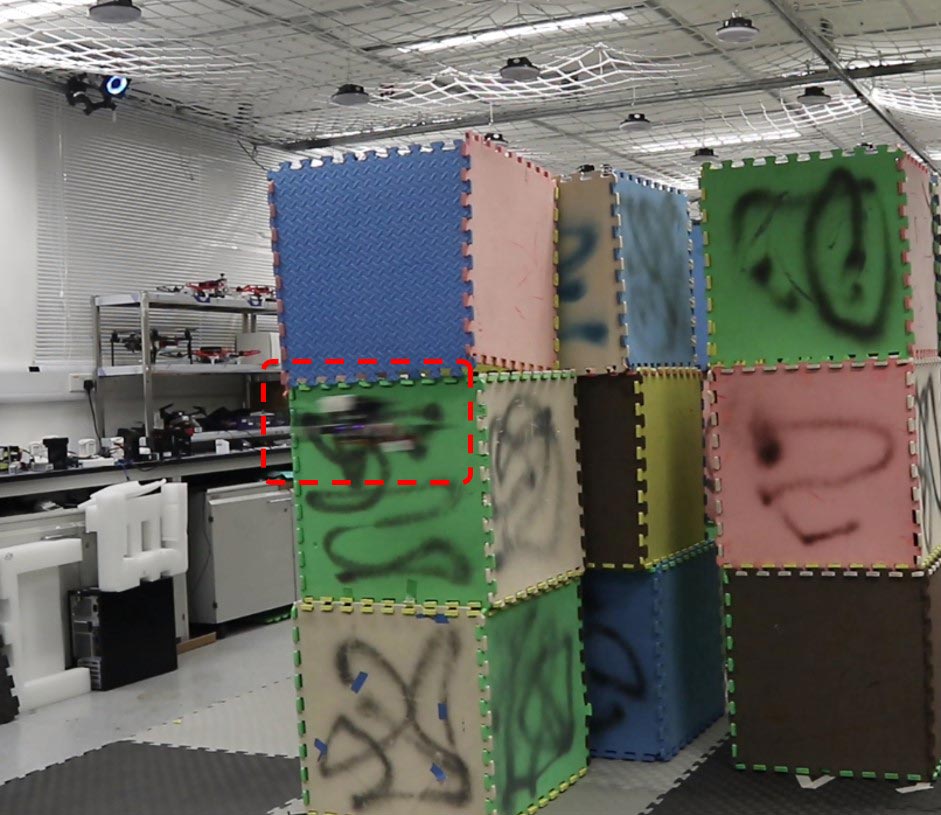}}       
		\subfigure[\label{fig:scene2_4}]
      {\includegraphics[width=0.485\columnwidth]{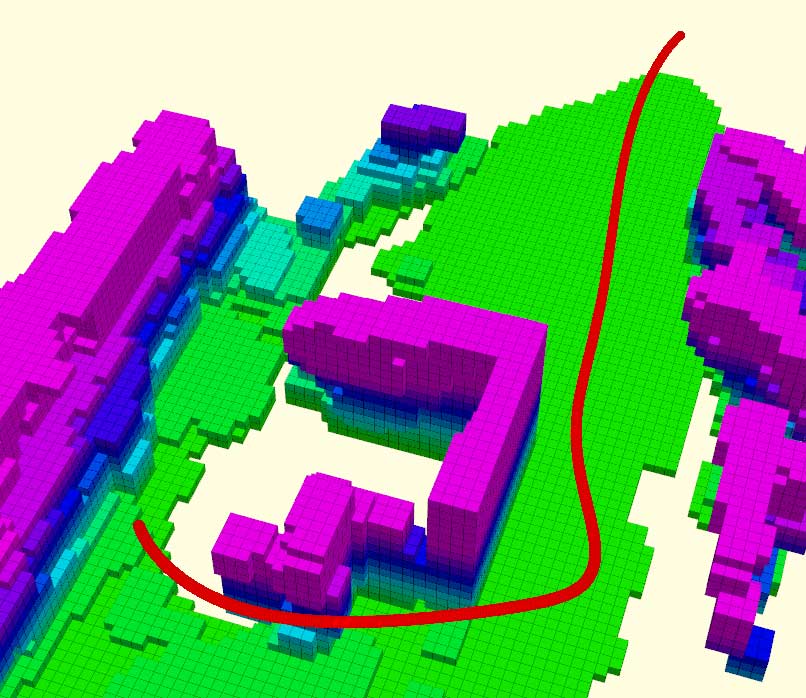}}       
      \vspace{-0.5cm}
	\end{center}
   \caption{\label{fig:scene2} Tests in scene 2.
   (a),(b) The quadrotor fails to react to the obstacle after turning right and crashes.
   (c),(d) The quadrotor avoid the obstacle behind the corner successfully.
   }
   \vspace{-0.1cm}
\end{figure*} 

\begin{figure}[t]
	\begin{center}          
		\subfigure[\label{fig:comp_refine_wo2}]
		{\includegraphics[width=0.495\columnwidth]{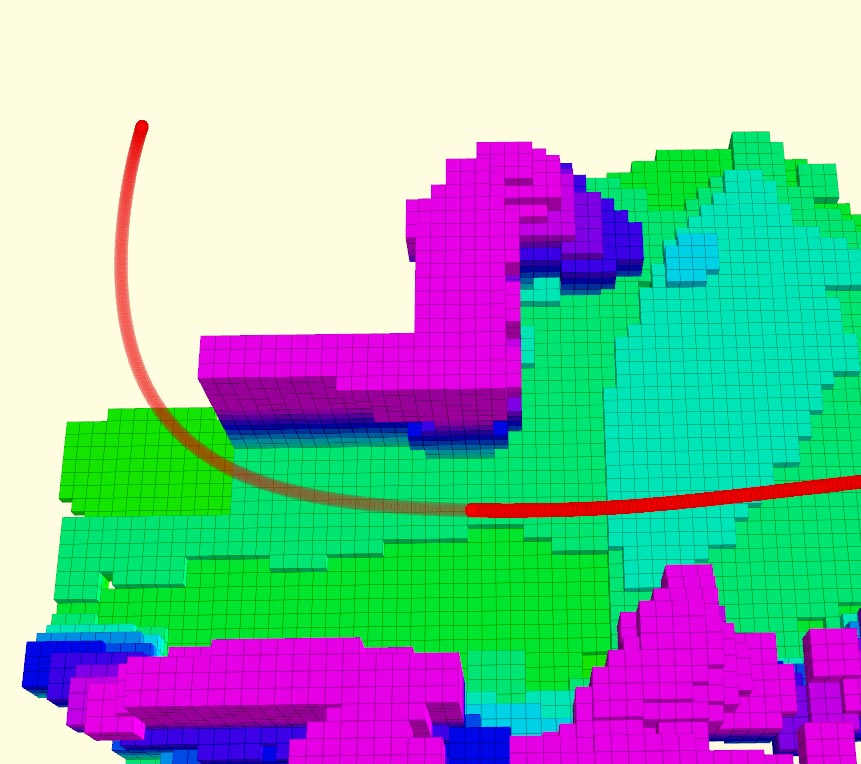}}       
		\subfigure[\label{fig:comp_refine_w2}]
      {\includegraphics[width=0.485\columnwidth]{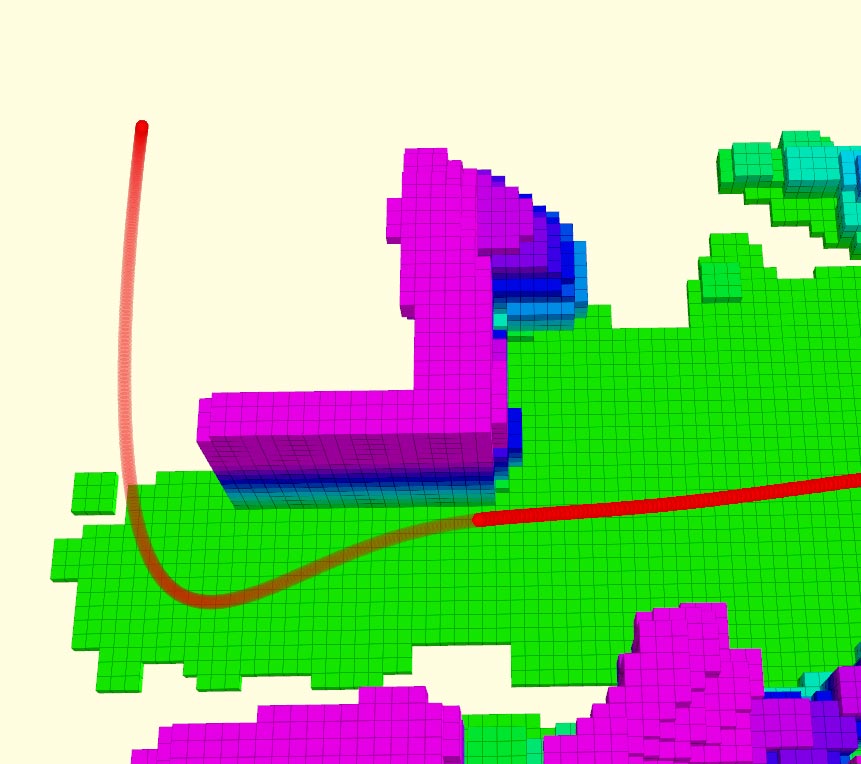}}       
      \vspace{-0.8cm}
	\end{center}
   \caption{\label{fig:comp_refine2} A comparison of the trajectories in scene 2 with (a) optimistic assumption and (b) risk-aware refinement.
   (a) The trajectory is close to the corner.
   (b) The trajectory is slightly longer around the corner, where the quadrotor can observe the unknown space in the back and react to possible dangers. 
   }
   \vspace{-0.1cm}
\end{figure} 

\begin{figure}[t]
	\begin{center}          
      {\includegraphics[width=0.49\columnwidth]{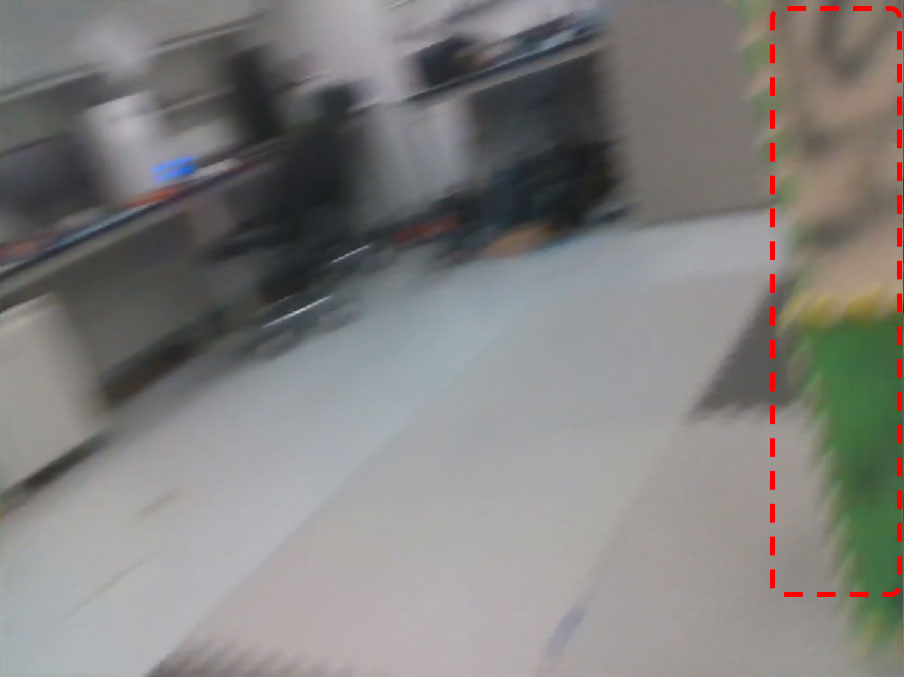}}       
      {\includegraphics[width=0.49\columnwidth]{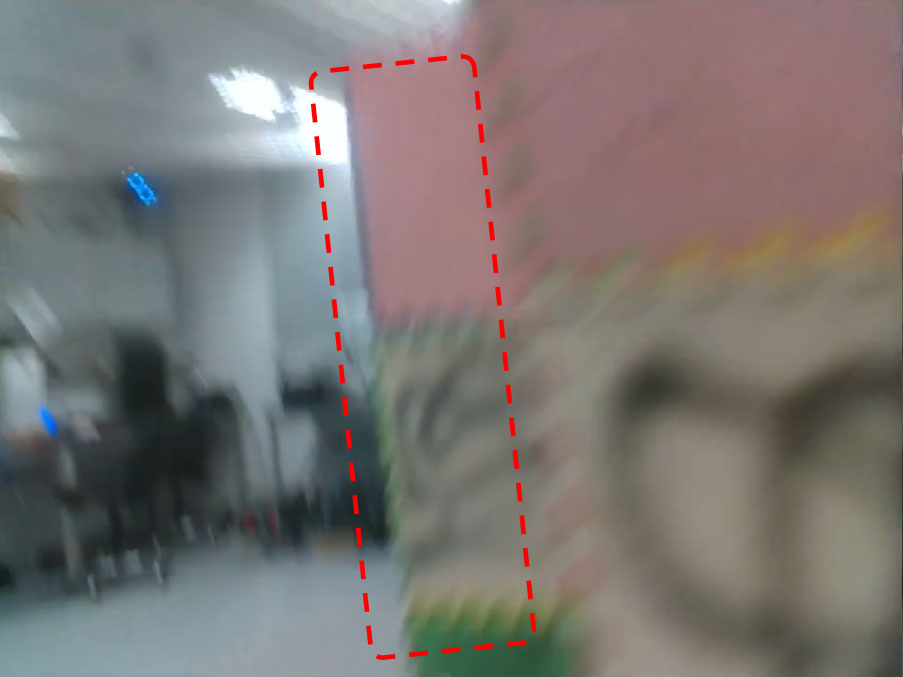}}       
		\subfigure[\label{fig:comp_yaw_2_1}]
		{\includegraphics[width=0.492\columnwidth]{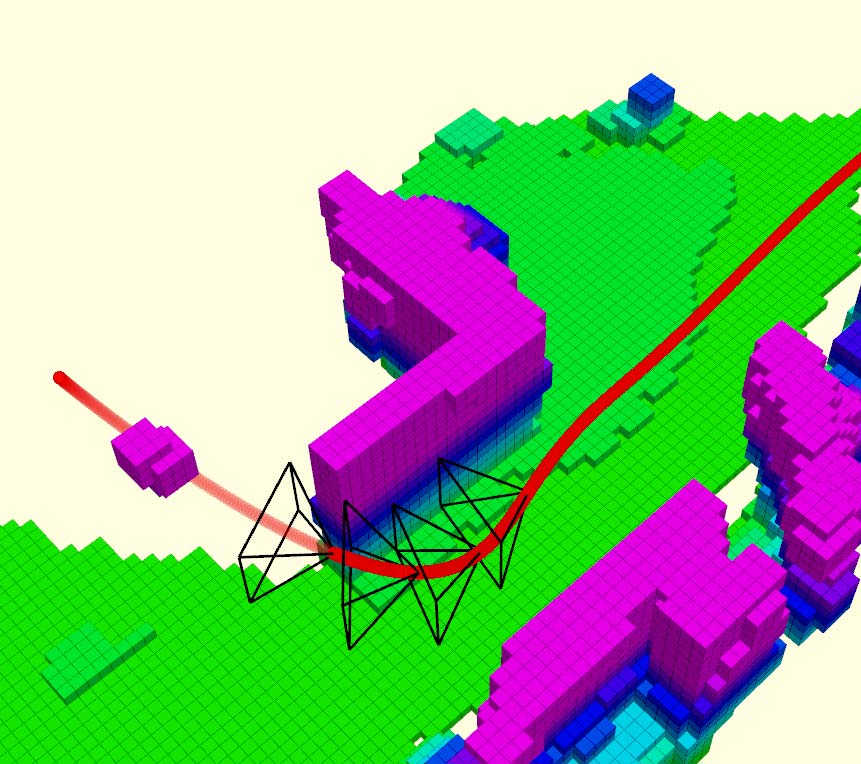}}       
		\subfigure[\label{fig:comp_yaw_2_2}]
      {\includegraphics[width=0.488\columnwidth]{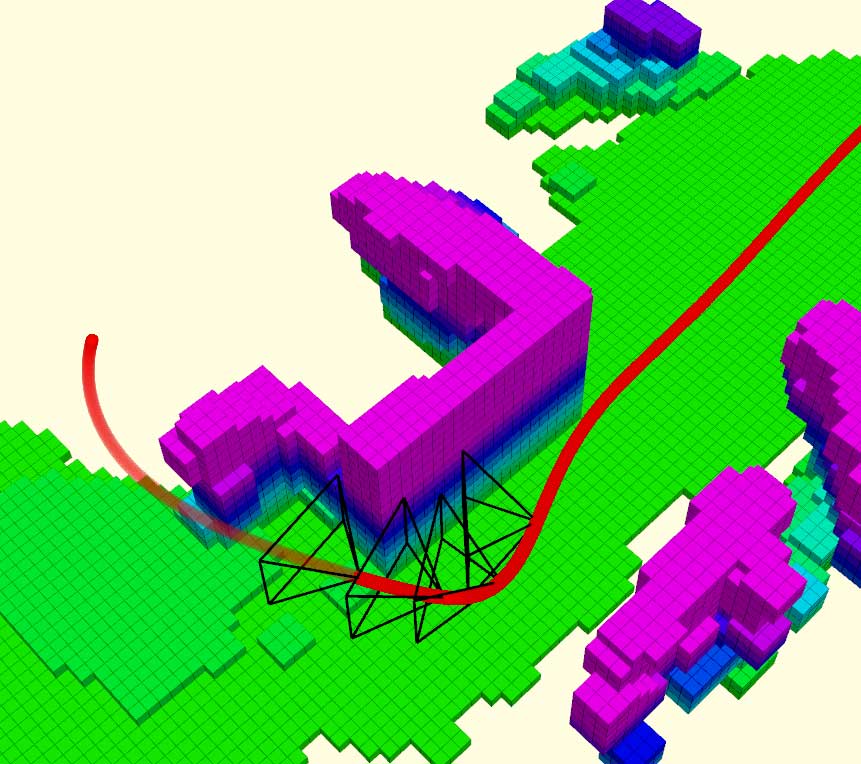}}       
      \vspace{-0.8cm}
	\end{center}
   \caption{\label{fig:comp_yaw_2} A comparison in scene 2 of (a) velocity-tracking yaw and (b) active exploration yaw. 
   (a) The quadrotor turns its head to the right later. The hidden obstacles is not observed until getting close.
   (b) The quadrotor looks to the right side earlier and discovers the obstacle behind the corner. 
   }
   \vspace{-0.5cm}
\end{figure}

\section{Results}
\label{sec:result}

\subsection{Implementation Details}
\label{subsec:implement}

We present tests in both real world and simulation.
In real-world experiments, a customized quadrotor platform equipped with an Intel RealSense Depth Camera D435 is used. 
All the state estimation, mapping, planning and control modules run on an Intel Core i7-8550U CPU.
For simulation, we use a simulating tool containing the quadrotor dynamics model, random map generator and depth image renderer.
The dynamics model relies on a numeric ODE solver \textit{odeint}\footnote{www.boost.org/doc/libs/1\_73\_0/libs/numeric/odeint/doc/html/index.html}.
The depth images are rendered in GPU by projecting point cloud of the surrounding obstacles onto the image plane.
Random noises are added to them to better mimic the real measurements. 
All simulations run on an Intel Core i7-8700K CPU and GeForce GTX 1080 Ti GPU.
The trajectory optimization is solved by a general non-linear optimization solver NLopt\footnote{https://nlopt.readthedocs.io/en/latest/}. 

\begin{table}[t]
   \centering
   \caption{\label{tab:real1} Four planners tested in real-world experiments.}
   \begin{tabular}{cccc} 
   \hline\hline
    \textbf{Planner}    &  \textbf{Strategy}    & \multicolumn{2}{c}{\textbf{Success Number}}  \\
    \cline{3-4}
   \multicolumn{1}{c}{} & \multicolumn{1}{c}{}  & \textbf{Scene 1} & \textbf{Scene 2}          \\ 
   
   A                    & Optimistic \& Velocity-tracking & 0                & 0                         \\
   B                    & Optimistic \& Active exploration & 0                & 0                         \\
   C                    & Risk-aware \& Velocity-tracking & 2                & 0                         \\
   D                    & Risk-aware \& Active exploration & \textbf{3}                & \textbf{3}                         \\
   \hline\hline
   \end{tabular}
   \vspace{-1.0cm}
\end{table}

\subsubsection{Global Planning}

We use the approach \cite{RicBryRoy1312} to compute global reference trajectories.
Note that we focus on evaluating the local replanning system, therefore, naive global trajectories are given, such as straight-line trajectory connecting the start and goal positions.

\subsubsection{Volumetric Mapping}
In all tests, the quadrotor starts with no prior knowledge of the environments. 
A volumetric mapping framework \cite{han2019fiesta} fuses the depth images from the stereo camera into a occupancy grid map. 
An ESDF is derived from the occupancy grid map using an efficient distance transform algorithm \cite{felzenszwalb2012distance} to support the gradient-based optimization (Sect.\ref{sec:pgo}) and visibility evaluation (Sect.\ref{sssec:visibmetric}).
Trilinear interpolation is also used to reduce the distance error induced by the discrete grid map.

\subsubsection{State Estimation and Control}
 
We localize the drone by a robust visual-inertial state estimator \cite{qin2017vins} in real-world tests.
In simulation, ground truth odometry is generated by the quadrotor dynamics model.
We use a geometric controller\cite{lee2010geometric} to track both the position and yaw trajectory. 

The following evaluation is divided into two parts. 
The first part evaluates the perception-aware planning strategy, the second part tests the whole replanning framework.

\begin{figure}[t]
   \begin{center}          
      \subfigure[\label{fig:bench2_data1} Avg. flight distance ($m$).]
      {\includegraphics[width=0.49\columnwidth]{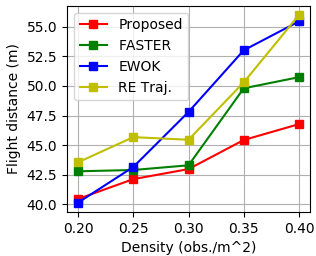}}       
      \subfigure[\label{fig:bench2_data2} Avg. flight time ($s$).]
      {\includegraphics[width=0.49\columnwidth]{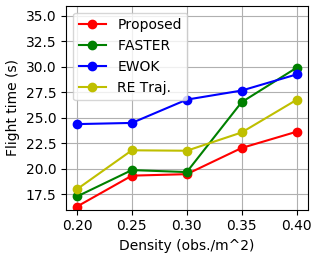}}       
      \subfigure[\label{fig:bench2_data3} Avg. consumed energy ($m^2/s^5$).]
      {\includegraphics[width=0.49\columnwidth]{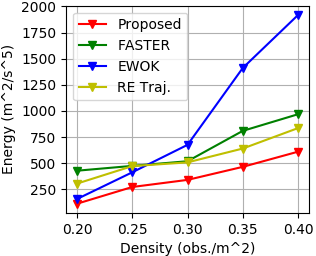}}       
      \subfigure[\label{fig:bench2_data4} Avg. replan time ($ms$).]
      {\includegraphics[width=0.49\columnwidth]{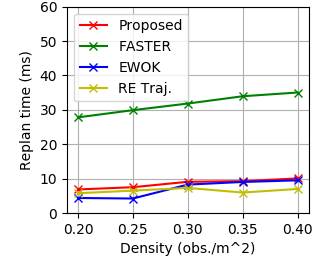}}       
      \subfigure[\label{fig:bench2_data6} Number of successful flights.]
      {\includegraphics[width=0.8\columnwidth]{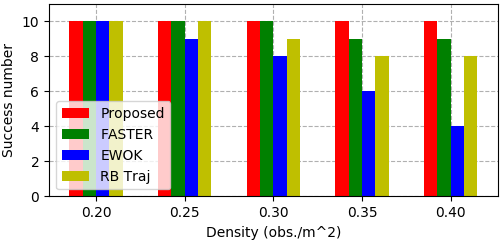}}       
      \vspace{-0.1cm}
   \end{center}
   \caption{\label{fig:bench2_data} Results of the comparisons between the proposed method and FASTER, EWOK and RE Traj..
   }
   \vspace{-1.0cm}
\end{figure} 

\begin{figure}[t]
   \begin{center}          
      \subfigure[\label{fig:bench_map1} Sample map 1: $0.25 \ \text{obs.}/m^2$]
		{\includegraphics[width=0.49\columnwidth]{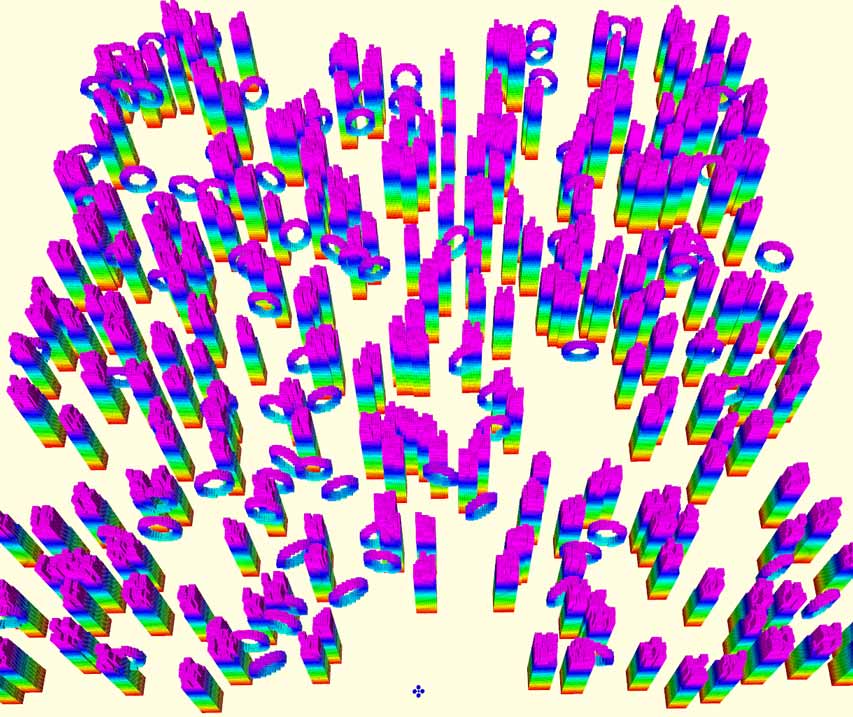}}       
      \subfigure[\label{fig:bench_map2} Sample map 2: $0.4 \ \text{obs.}/m^2$]
      {\includegraphics[width=0.49 \columnwidth]{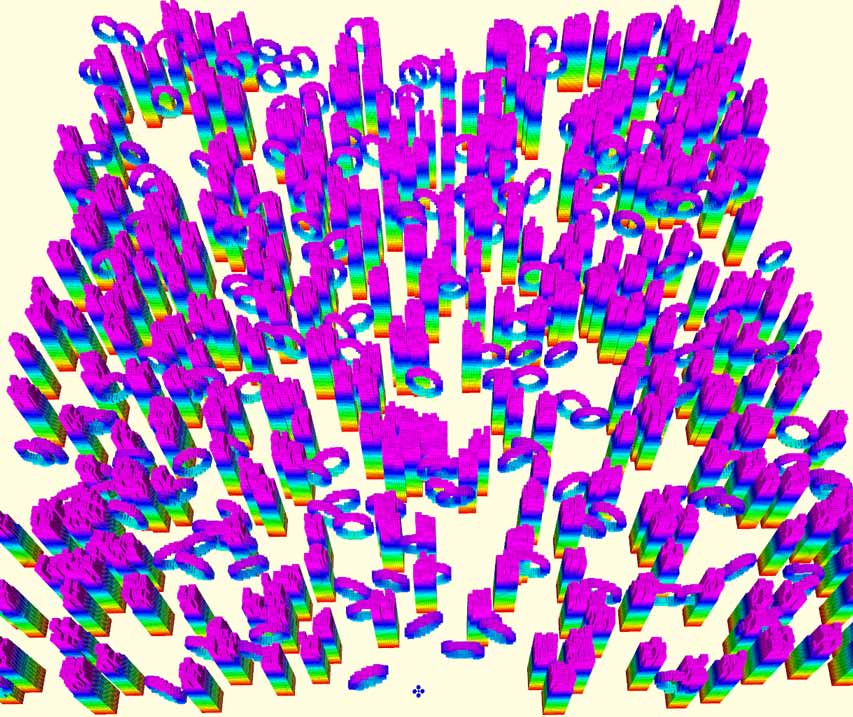}}       
      \subfigure[Trajectories generated by the proposed method (red), FASTER\cite{tordesillas2019faster} (green), EWOK\cite{usenko2017real} (cyan) and RE Traj.\cite{zhou2019robust} (yellow).
      Obstacles are set as gray transparent for clarity. \label{fig:bench_traj}]
      {\includegraphics[width=0.98\columnwidth]{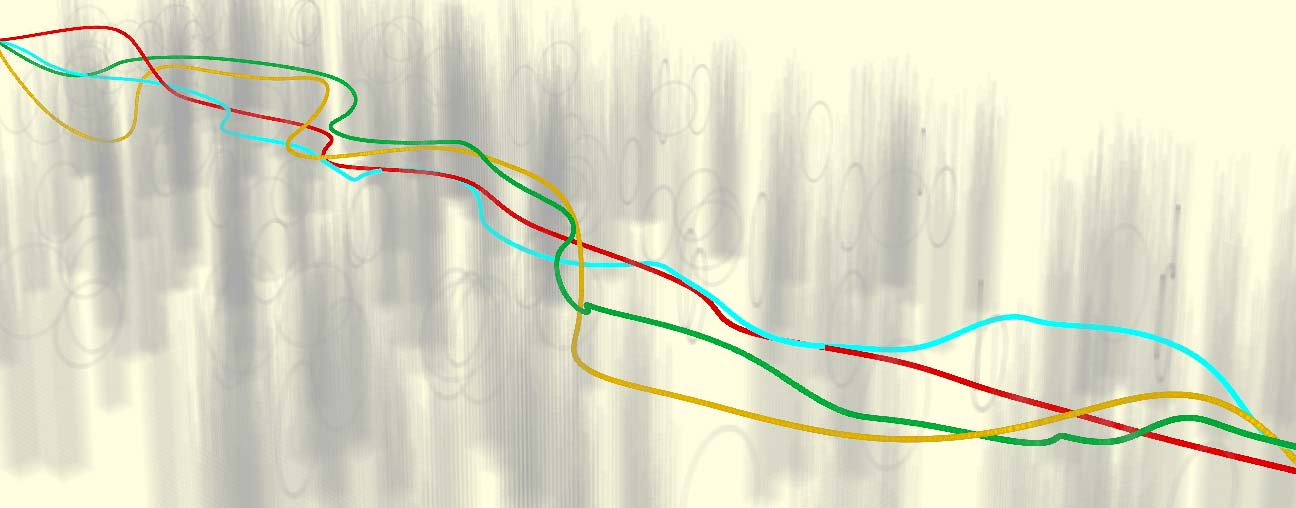}}       
      \vspace{-0.1cm}
   \end{center}
   \caption{\label{fig:bench2} Samples of the maps and generated trajectories.
   }
   \vspace{-0.5cm}
\end{figure}

\subsection{Perception-aware Planning Strategy}

\subsubsection{Real-world Tests}
\label{subsubsect:real1}

We conduct comparative experiments to show the importance of introducing active perception.
Specifically we compare the proposed strategy: the risk-aware refinement (Sect.\ref{sec:refinement}) and active exploration yaw (Sect.\ref{sec:yaw}) with the commonly used ones: the optimistic assumption and the velocity-tracking yaw.
Optimistic assumption treats all unknown space as collision-free, which is frequently adopted such as in \cite{fei2018jfr,liu2017planning, zhou2019robust, usenko2017real}.
The velocity-tracking yaw relates the desired yaw angle to the velocity: $\phi(t) = \text{arctan} (\frac{v_y(t)}{v_x(t)}) $, to increase the chance of seeing obstacles.
Four local planners listed in Tab.\ref{tab:real1} are tested in two scenes.
Each planner is tested 3 times in both scenes and we record the number of successful flights.
The maximum velocity and acceleration are set as $ 3m/s$ and $ 2.5m/s^2$.
Whenever collision along the trajectory is detected and the collision point is closer than $0.5m$, emergency stop is conducted immediately for safety.

In the first scene, a straight-line global reference trajectory is given (Fig.\ref{fig:active_scene_1}).
A large obstacle consisting of several boxes and boards are placed on the way.
When the quadrotor approaches the obstacle, boxes in the front row will be revealed first, while others behind them are occluded and invisible at the beginning. 
Planners with optimistic assumption (A \& B) are unaware of the potential danger behind. 
They simply replan trajectories to avoid the viewed boxes, along which there is low visibility to the boxes in the back, as showed in Fig.\ref{fig:comp_refine_wo}.
As a result, the quadrotor gets 'surprised' by the occluded boxes afterwards and pauses in emergency, as showed in Fig.\ref{fig:4scene_1} and \ref{fig:4scene_2}. 
In contrast, planners with risk awareness (C \& D) generate trajectories that deviate a bit more laterally, along which visibility toward the unknown area in the back is higher (Fig.\ref{fig:comp_refine_w}).
Therefore in both cases the quadrotor reach the goal more times. 
However, with the velocity-tracking yaw (planner C), the quadrotor does not face toward the unknown area in the back quickly, which postpones the discovery of occluded boxes and causes 1 failure.
In comparison, with the active exploration yaw (planner D) the quadrotor quickly turns toward the unknown area and observes the previously occluded boxes, enabling itself to take action earlier. 
This comparison is displayed in Fig.\ref{fig:comp_yaw}.

In the second scene, an obstacle is placed right behind the corner, which is invisible to the quadrotor until it turns right.
The reference trajectory is set to pass through the obstacles deliberately (Fig.\ref{fig:active_scene_2}).
In this scene, the quadrotor can only reach the goal safely with planner D.
For other three planners, the quadrotor collides with the obstacle behind the corner, due to either the poor visibility of the replanned trajectories (planner A \& B), or the delay of perception caused by the velocity-tracking yaw (planner C).
Note that even emergency stop is conducted, the quadrotor fail to avoid collision in time (Fig.\ref{fig:scene2}).
The comparisons of the replanned trajectories and yaw angle are displayed in Fig.\ref{fig:comp_refine2} and Fig.\ref{fig:comp_yaw_2} respectively.

The experiments demonstrate two critical factors to survive in high-speed flights:
(a) having good visibility toward the unknown regions that will influence the flight and 
(b) looking toward the relevant direction to eliminate those unknown regions actively.
The proposed method takes into account these factors and guarantees safety for fast flight.
More details about the experiments are presented in the attached video.

\subsubsection{Benchmark Comparisons}
\label{subsubsect:bench1}

We compare the proposed strategy with the safe local exploration (SLE) presented in \cite{oleynikova2018safe} in simulation.
This strategy originated from the "next-best-view" planner \cite{bircher2018receding} in the exploration literature, but is adapted for online functioning in goal reaching tasks.
It repeatedly selects intermediate goals that are closer to the final goal and have higher information gain, after which a local planner replans new trajectories toward the goals. 
It also adopts the velocity-tracking yaw, as is detailed in Sect.\ref{subsubsect:real1}.
To compare the strategies fairly, we integrated both of them with our robust optimistic replanning (Sect.\ref{sec:pgo}, \ref{sec:topo_path}). 
They are tested in $10$ random maps with $5$ different obstacle densities, $5$ trials are conducted for each map.
We compare the number of successful flight, flight time and flight distance. 
Samples of the maps are displayed in Fig.\ref{fig:bench_map1}, \ref{fig:bench_map2}.

As is shown in Tab.\ref{tab:bench1}, the proposed strategy achieves higher number of successful flights when the scene gets more cluttered.
Our strategy enforces visibility to dangerous unknown areas, and control the yaw angle to observe those areas actively.
Therefore it can guarantee safety even the environment becomes very complex.
For SLE, the velocity-tracking yaw is the major cause of failure, as it may not face toward dangerous unknown regions in time, as has been shown in Sect.\ref{subsubsect:real1}. 

\begin{figure*}[t]
   \begin{center}          
      \subfigure[\label{fig:indoor1} The quadrotor passes a horizontal cardboard, after which it will avoid the vertical pillar.]
      {\includegraphics[width=0.65\columnwidth]{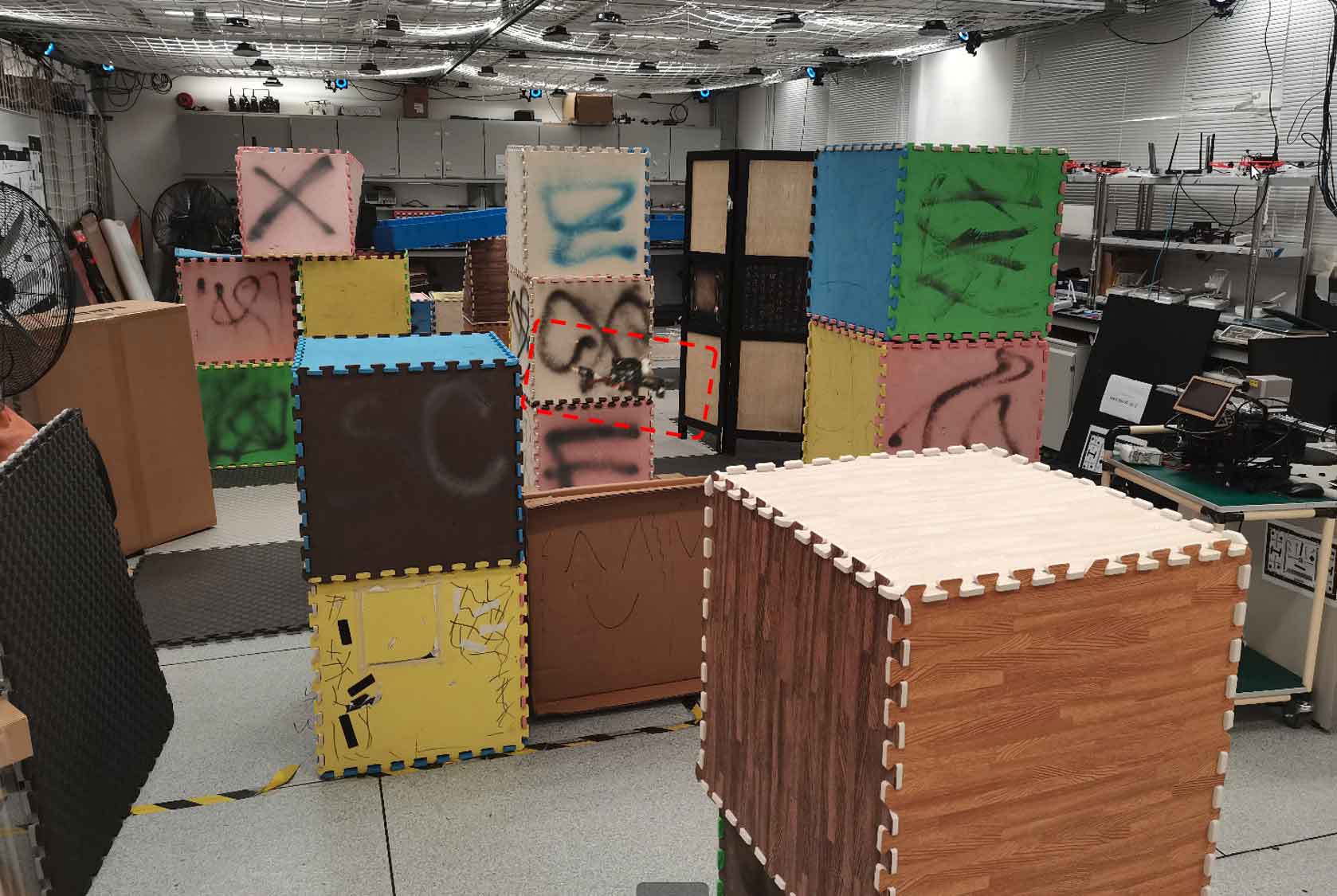}}       
      \subfigure[\label{fig:indoor2} Composite image of the flight experiment.]
      {\includegraphics[width=0.65\columnwidth]{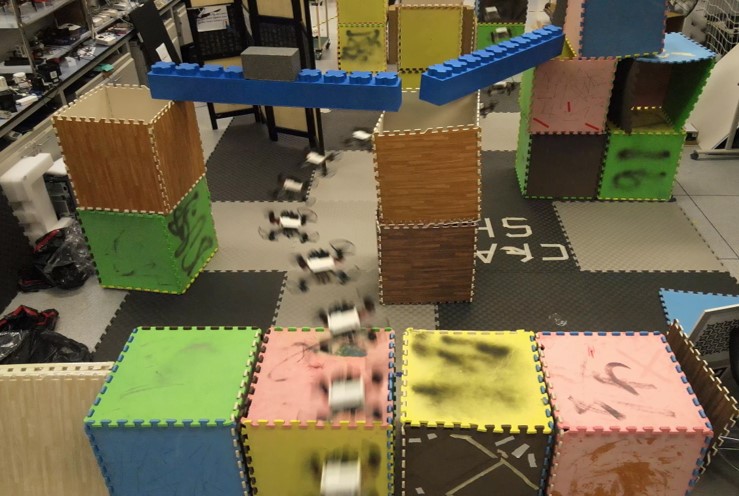}}       
      \subfigure[\label{fig:indoor3} The quadrotor flies in the narrow passages and avoids boxes.]
      {\includegraphics[width=0.65\columnwidth]{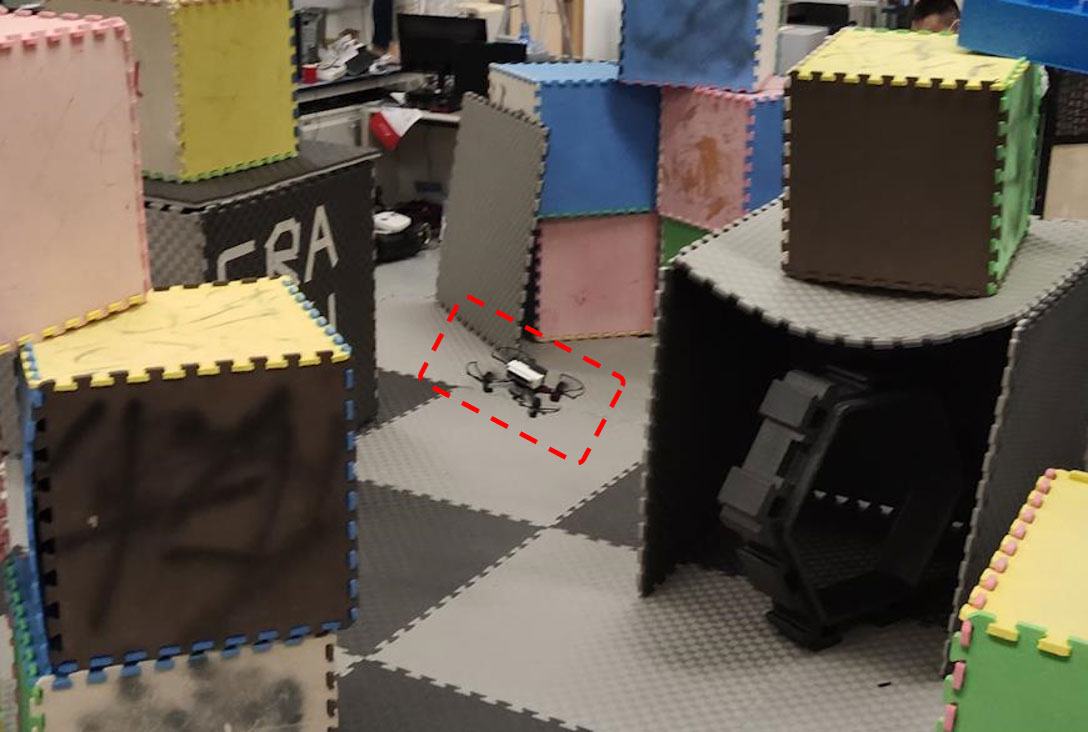}}       
      \vspace{-0.1cm}
   \end{center}
   \caption{\label{fig:indoor} Three indoor scenes for the fast flight experiments.
   Several tests are conducted for each scenes, more details are available in the attached video. 
   }
   \vspace{-0.3cm}
\end{figure*} 

\begin{figure}[t]
   \begin{center}          
      \subfigure[\label{fig:indoor_bag1} ]
      {\includegraphics[width=0.98\columnwidth]{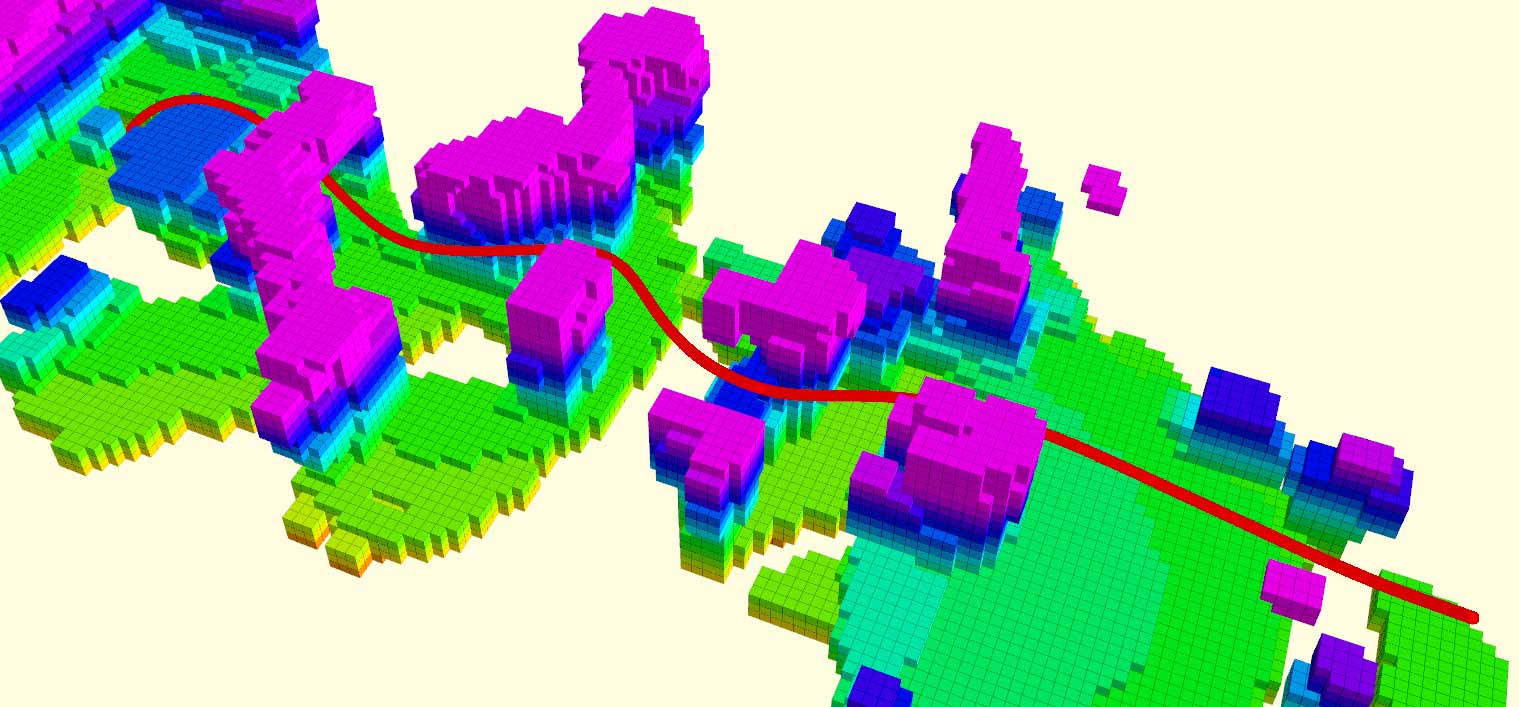}}       
      \subfigure[\label{fig:indoor_bag2} ]
      {\includegraphics[width=0.49\columnwidth]{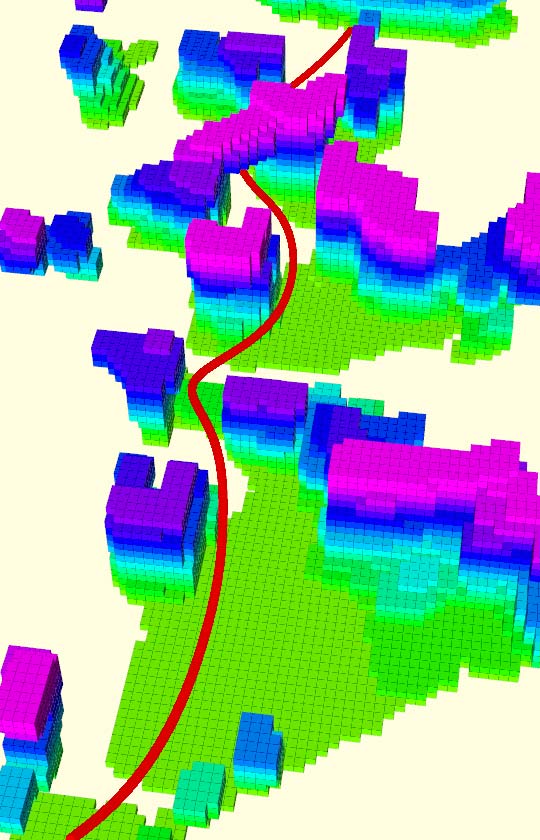}}       
      \subfigure[\label{fig:indoor_bag3} ]
      {\includegraphics[width=0.49\columnwidth]{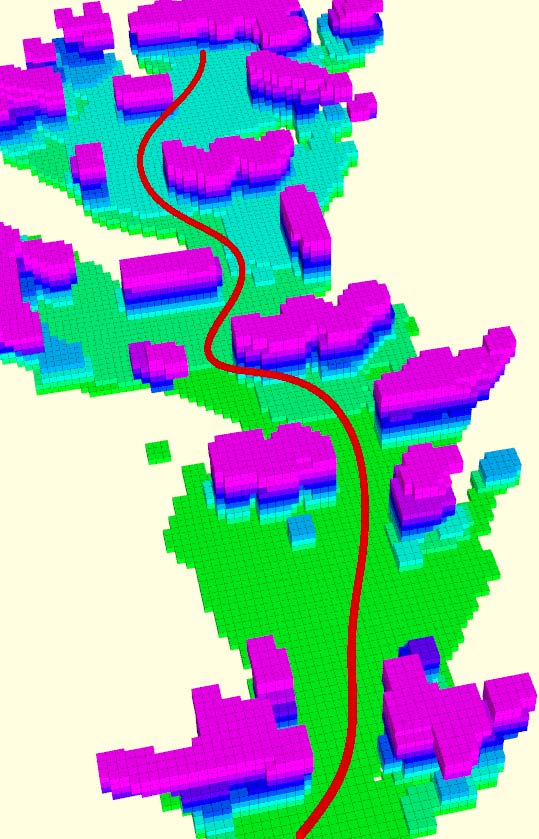}}       
      \vspace{-0.1cm}
   \end{center}
   \caption{\label{fig:indoor_bag} Samples of the online generated maps and trajectories executed by the quadrotor in the different indoor scenes.
   }
   \vspace{-0.6cm}
\end{figure}

\begin{figure}[t]
   \begin{center}          
      \subfigure[\label{fig:vel_in} ]
      {\includegraphics[width=0.85\columnwidth]{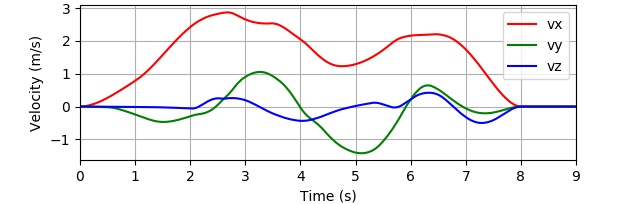}}       
      \subfigure[\label{fig:vel_out} ]
      {\includegraphics[width=0.999\columnwidth]{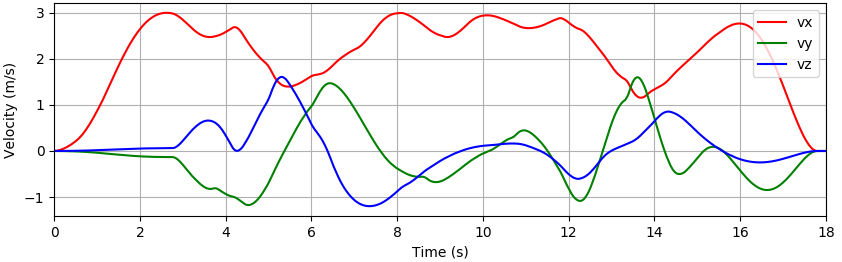}}       
      \vspace{-0.6cm}
   \end{center}
   \caption{\label{fig:vel_profile} Velocity profiles of sample flights in (a) indoor experiment (b) forest experiment.
   }
   \vspace{-0.5cm}
\end{figure} 

Besides, our strategy is also more beneficial to achieve lower flight distance and time than SLE. 
SLE only selects intermediate goals and plans within the known unoccupied space, which is conservative.
Besides, since the selection of intermediate goals takes information gain into account, the quadrotor tends to take some detours to gather more information, which leads to longer flight distance and time.
In contrast, our strategy plans in both the known and unknown space, allowing more aggressive behaviors under the premise of safety.
Moreover, instead of treating all unknown areas equally, it only focus on observing areas that are more relevant to the flight, which eliminates many unnecessary detours and improve the overall flight efficiency.


\subsection{Replanning Framework for Fast Flight}

\subsubsection{Benchmark Comparisons}

We compare our replanning framework with several state-of-the-art methods, FASTER \cite{tordesillas2019faster}, EWOK\cite{usenko2017real} and RE Traj.\cite{zhou2019robust}.
FASTER belongs to the hard-constrained category (Sect.\ref{subs:related_trajectory}), and features maintaining a feasible and safe back-up trajectory in the free-known space at each replanning step to improve safety.
It also adopts a mixed integer quadratic program (MIQP) formulation to obtain a more reasonable time allocation of the trajectories.
Both \cite{usenko2017real,zhou2019robust} belong to the gradient-based methods.
They utilize a uniform B-spline trajectory representation to replan efficiently.
\cite{zhou2019robust} further exploits the convex hull property of B-spline and introduces a kinodynamic path searching to find more promising initial trajectories.
We also test the four methods in 10 random maps with 5 obstacle densities.
Note that all benchmarked methods are open-source and we use their default parameter settings.
The number of successful flights, average flight distance, flight time, energy (integral of squared jerk), computation time of each replanning, and total replan number in each flight are recorded. 
Samples of the maps and the trajectories generated by the four methods are shown in Fig.\ref{fig:bench2}.

\begin{table}[t]
   \centering
   \caption{Comparisons of the perception-aware planning strategies. \label{tab:bench1}}
   \begin{tabular}{ccccc} 
   \hline\hline
   \begin{tabular}[c]{@{}c@{}}\textbf{Density} \\ (obs./$m^2$) \end{tabular}                 & \textbf{Method}  & \begin{tabular}[c]{@{}c@{}}\textbf{Number of} \\\textbf{Success} \end{tabular} & \begin{tabular}[c]{@{}c@{}}\textbf{Flight} \\\textbf{Distance (m)}\end{tabular} & \begin{tabular}[c]{@{}c@{}}\textbf{Flight}\\ \textbf{Time (s)}\end{tabular}  \\ 
   \hline
   \multirow{2}{*}{0.2}     & SLE\cite{oleynikova2018safe}      & 10                                                         & 43.475                                                        & 20.012                                                           \\ 
                             & Proposed                         & 10                                                  & \textbf{41.822}                                                         & \textbf{16.608}                                                           \\ 
   \hline
   \multirow{2}{*}{0.25}& SLE\cite{oleynikova2018safe}        & 10                                                         & 44.682                                                          & 20.593                                                        \\ 
                             & Proposed                      & 10                                                      & \textbf{42.133}                                                         & \textbf{19.338}                                                           \\ 
   \hline
   \multirow{2}{*}{0.3} & SLE\cite{oleynikova2018safe}         & 9                                                         & 44.874                                                          & 22.316                                                        \\ 
                             & Proposed                      & \textbf{10}                                                     & \textbf{42.982}                                                         & \textbf{19.483}                                                           \\
   \hline
   \multirow{2}{*}{0.35} & SLE\cite{oleynikova2018safe}         & 9                                                         & 46.093                                                          & 23.577                                                        \\ 
   & Proposed                                             & \textbf{10}                                                       & \textbf{45.436}                                                         & \textbf{22.050}                                                           \\
   \hline
   \multirow{2}{*}{0.4} & SLE\cite{oleynikova2018safe}         & 8                                                         & 47.817                                                          & 27.903                                                         \\ 
   & Proposed                      & \textbf{10}                                                     & \textbf{46.776}                                                         & \textbf{23.639}                                                           \\
   \hline\hline
   \end{tabular}
   \vspace{-0.7cm}
\end{table}

As displayed in Fig.\ref{fig:bench2_data}, our method outperform others in the aspects of flight distance, flight time and energy consumption, with competitive computation efficiency.  
FASTER rarely fails in the tests, thanks to the back-up trajectories. 
However, due to the computationally demanding MIQP formulation, its overhead is higher.
The other two benchmarked methods are more efficient. 
However, EWOK suffers from the local minima issue, so it usually fails or outputs low-quality solutions in dense environments.
The kinodynamic path searching and B-spline optimization adopted by RE Traj. relieve the local minima significantly.
Nonetheless, due to the lack of perception consideration, the succuss number is mediocre in dense environments. 
Compared to them, the proposed method search the solution space effectively with the guidance of topologically distinctive paths, and generates high-quality trajectories consistently.
Safety is also reenforced by introducing perception awareness.


\begin{figure*}[t]
   \begin{center}          
      \subfigure[\label{fig:outdoor1} Flying through forest 1.]
      {\includegraphics[width=0.65\columnwidth]{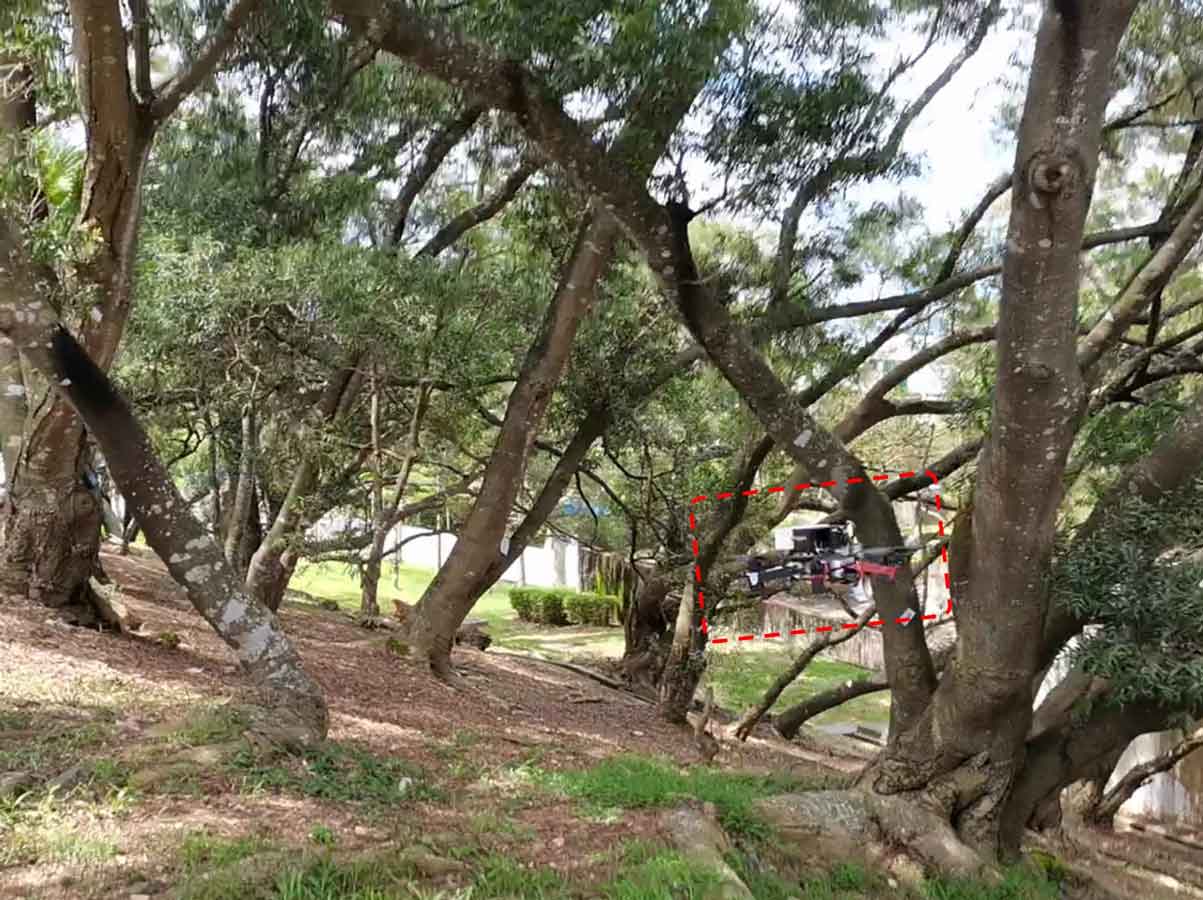}}       
      \subfigure[\label{fig:outdoor2} Flying up the slope in forest 1.]
      {\includegraphics[width=0.65\columnwidth]{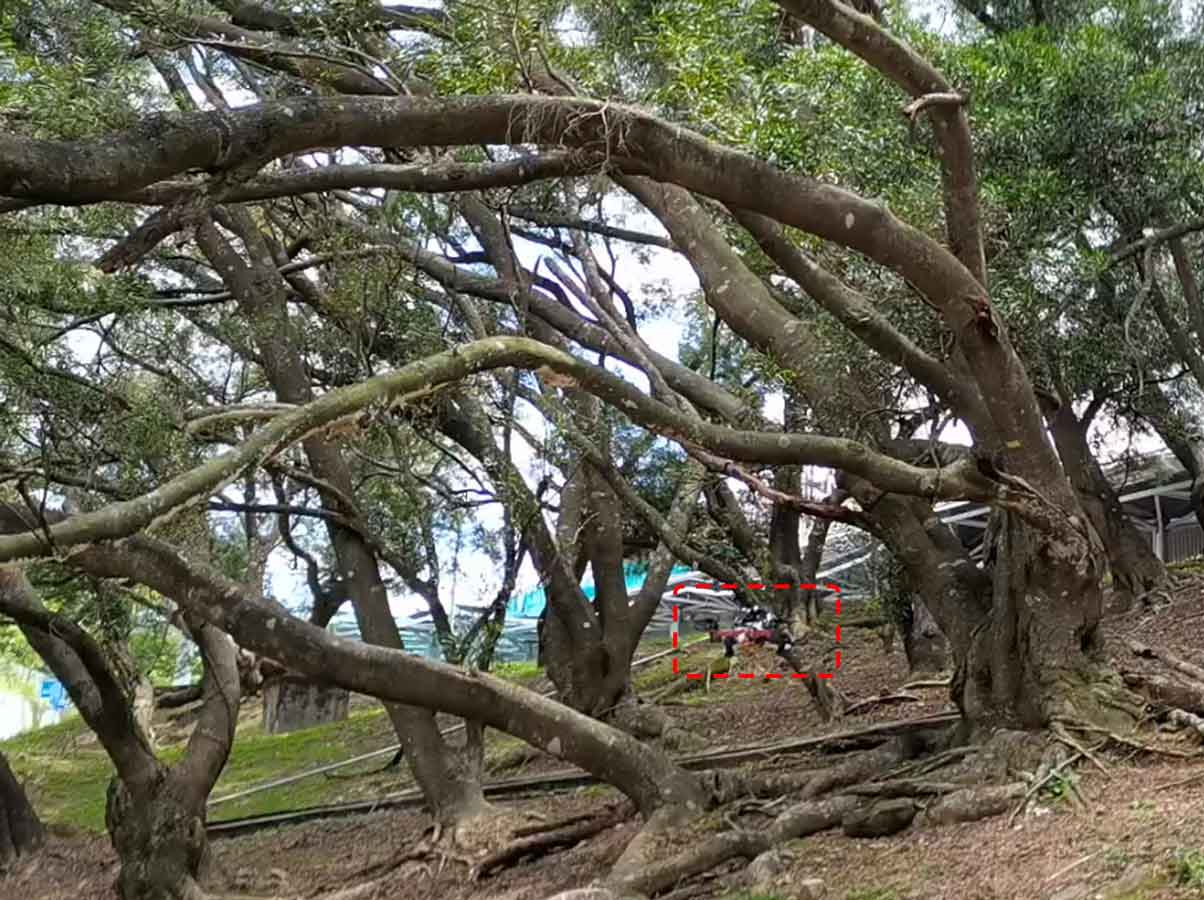}}       
      \subfigure[\label{fig:outdoor3} Flying through forest 2.]
      {\includegraphics[width=0.65\columnwidth]{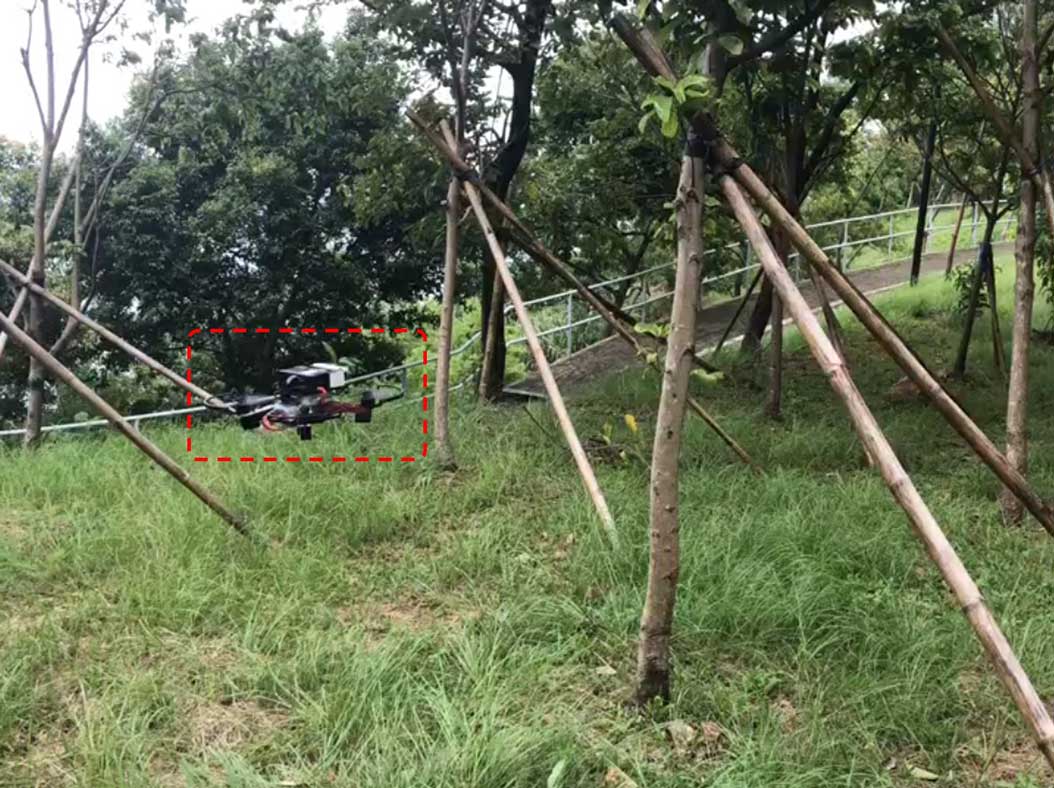}}       
      \vspace{-0.1cm}
   \end{center}
   \caption{\label{fig:outdoor} Autonomous fast flight experiments in two dense forests.
   }
   \vspace{+0.2cm}
\end{figure*} 

\begin{figure*}[t]
   \begin{center}          
      \subfigure[\label{fig:outdoor_bag1} Flight in forest 1.]
      {\includegraphics[width=1.99\columnwidth]{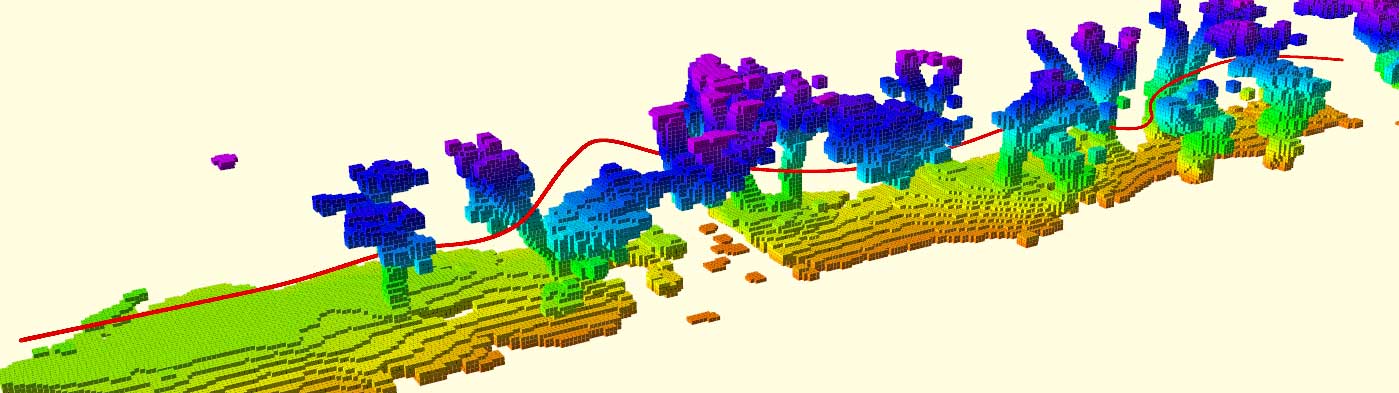}}       
      \subfigure[\label{fig:outdoor_bag2} The quadrotor flies up the slope to the first goal (green circle), after which it flies toward the second goal.]
      {\includegraphics[width=0.999\columnwidth]{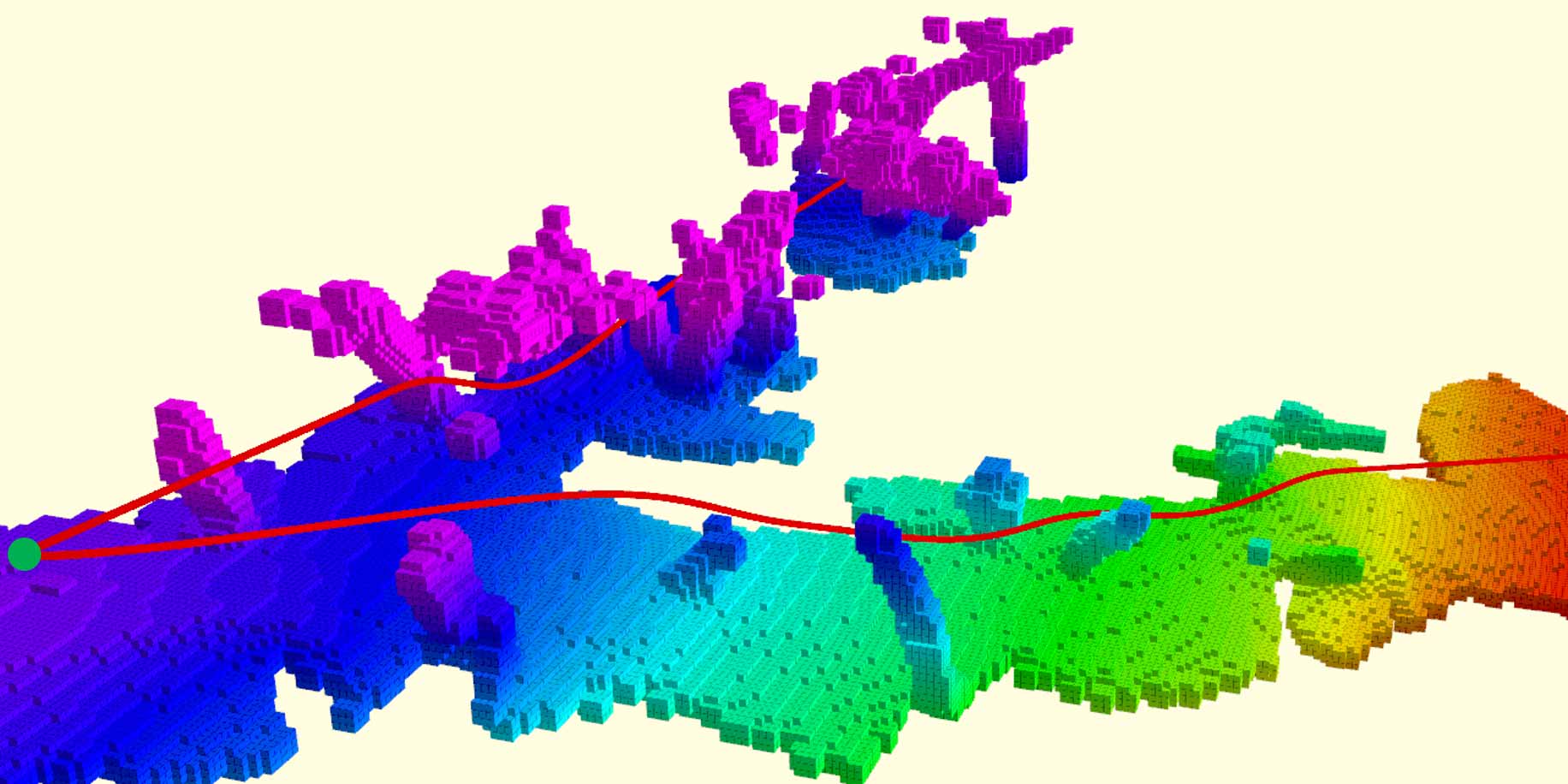}}       
      \subfigure[\label{fig:outdoor_bag3} Flight in forest 2.]
      {\includegraphics[width=0.98\columnwidth]{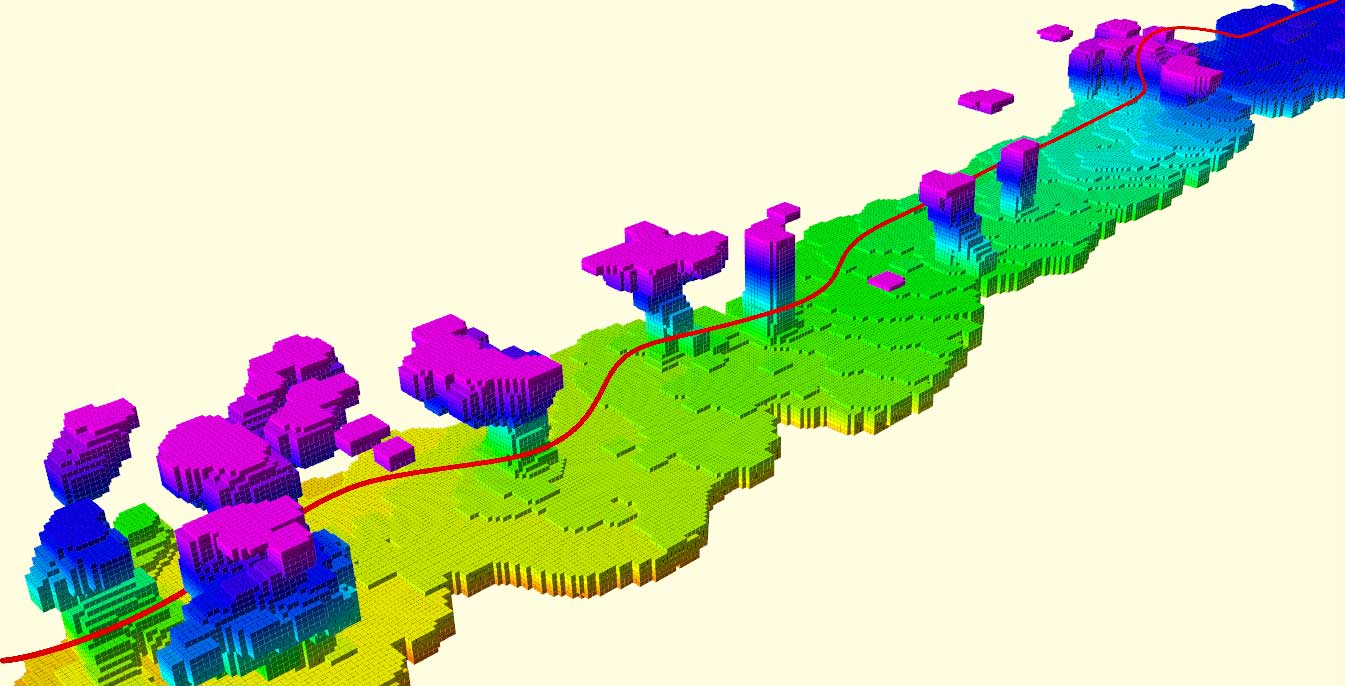}}       
      \vspace{-0.1cm}
   \end{center}
   \caption{\label{fig:outdoor_bag} Online generated maps of the forests and trajectories executed by the quadrotor.
   The map is colored by height.
   }
   \vspace{+0.2cm}
\end{figure*} 

\subsubsection{Indoor Flight Test}

We conduct aggressive flight experiments in three indoor scenes (Fig.\ref{fig:flight}, \ref{fig:indoor}) to validate our planning system.
Various types of obstacles are placed randomly and densely to make up the challenging flight environments. 
Distance of neighboring obstacles are only around 1 meter, making the space for safe navigation very limited.
Besides, the high obstacle density makes visibility to the environment very restricted, since many obstacles are occluded by others, which poses greater challenges to the replanning algorithm.

In each experiment, the final goal is set to 14$m$ away from the quadrotor.
Straight-line global reference trajectories are given and local replanning is conducted within a horizon of 7 $m$. 
Samples of the online generated map and executed trajectories are presented in Fig.\ref{fig:indoor_bag}.
The velocity profile of one flight are showed in Fig.\ref{fig:vel_in}, in which the maximum speed is $2.90 \ m/s$ and average speed is $1.77 \ m/s$.
The flight distance and time are $ 14.12 \ m$ and $ 8.0 \ s$ respectively.
We refer the readers to the attached video for more tests.

\subsubsection{Outdoor Flight Test}

Finally, we conduct fast flight tests in three different outdoor scenes, as displayed in in Fig.\ref{fig:outdoor}, to validate our planning method in natural environments. 
The outdoor environments are typically unstructured and irregular, where the quadrotor should perform agile 3D maneuvers to avoid obstacles such as rocks and branches and leaves of trees.
Note that despite the outdoor environments, we do not use external devices for localization.

Results of the online generated map and executed trajectories are presented in Fig.\ref{fig:outdoor_bag}.
In the first scene, the quadrotor flies through the forest to the goal $39 \ m$ away from the initial position.
The velocity profile is showed in Fig.\ref{fig:vel_out}.
The maximum speed is $3.19 \ m/s$ and average speed is $ 2.29 \ m/s $.
The flight takes $40.78 \ m$ and $17.83 \ s$.
In the second scene, the quadrotor flies up a slope to the first goal, after which it flies to the second goal. 
The first goal is $ 30 \ m $ away and the change in height is $ 7 \ m $.
The second goal is $ 17 \ m $ far.
The whole flight takes $ 48.71 \ m $ and $ 23.19 \ s $.
The third scene is a larger forest, where the goal is set to $45 \ m$ away.
The flight distance is $ 46.85 \ m $, which takes $ 21.91 \ s $ to finish.  
The maximum and average speed are $ 3.41 \ m/s $ and $ 2.14 \ m/s $.
More details of the flights are showed in the attached video.

\vspace{+0.5cm}
\section{Conclusions}
\label{sec:conclude}

In this paper, we propose a robust and perception-aware replanning method for high-speed quadrotor autonomous navigation. 
The path-guided optimization and topological path searching are devised to escape from local minima and explore the solution space more thoroughly, through which higher robustness and optimality guarantee are obtained.
The robust planner is further enhanced by the perception-aware strategy, which takes special caution about regions that may be dangerous to the quadrotor.
The yaw angle of the quadrotor is also planned to actively explore the environments, especially areas that are relevant to the future flight.
The planning system is evaluated comprehensively through benchmark comparisons. 
We integrate the planning method with global planning, state estimation, mapping, and control into a quadrotor platform and conduct extensive challenging indoor and outdoor flight tests.
Results show that the proposed method is robust and capable of supporting fast and safe flights.  
We release the implementation of our system to the community.

\addtolength{\textheight}{0cm}   

\newlength{\bibitemsep}\setlength{\bibitemsep}{0.2\baselineskip}
\newlength{\bibparskip}\setlength{\bibparskip}{0.95pt}
\let\oldthebibliography\thebibliography
\renewcommand\thebibliography[1]{%
\oldthebibliography{#1}%
\setlength{\parskip}{\bibitemsep}%
\setlength{\itemsep}{\bibparskip}%
}
\bibliography{zby} 


\end{document}